\documentclass[reprint,aps,pre]{revtex4-1}

\usepackage{amsmath,amsfonts,amscd,amssymb,graphicx,amsthm}

\theoremstyle{definition}

\usepackage{subeqnar}
\usepackage{dcolumn}
\usepackage{booktabs}
\usepackage{overpic}
\usepackage{color}
\usepackage[caption=false]{subfig}
\usepackage{adjustbox}
\usepackage{rotating}
\usepackage{float}
\makeatletter
\let\newfloat\newfloat@ltx
\makeatother
\usepackage{algorithm}
\usepackage{algorithmic}
\usepackage{placeins}
\usepackage{algorithm}
\usepackage{comment}
\usepackage{soul, xcolor}

\begin{document}

\title{Physics-enhanced reinforcement learning \\ for real-time optimal control of dynamical systems}

\author{Matteo Tomasetto$^\dag$, Nicolò Botteghi$^\ddag$, Gabriele Bruni$^\ddag$, and Andrea Manzoni$^\ddag$}
\affiliation{$^\dag$ Department of Mechanical Engineering, Politecnico di Milano, Milano, Italy}
\affiliation{$^{\ddag}$ MOX - Department of Mathematics, Politecnico di Milano, Milano, Italy}

\begin{abstract}
Reinforcement learning (RL) has recently emerged as a promising feedback control strategy for nonlinear and complex dynamical systems. However, RL algorithms are sample inefficient and require a large number of interaction with the environment to synthesize optimal control strategies. Consequently, applications of RL are typically limited to sparse sensors and actuators due to the curse of dimensionality entailed by the exploration-exploitation dilemma in high-dimensional spaces. In this work, we bridge RL and traditional optimal control for dynamical system with a novel {\em Physics-EnhAnced Reinforcement Learning} (PEARL) paradigm tailored to the control of high-dimensional and parametric dynamical systems, exploiting the differentibility of their dynamics. Specifically, PEARL employs an actor-adjoint algorithm that leverages automatic differentiation to compute policy gradients over short horizons and adjoint-based sensitivities of future returns approximated via neural networks, significantly reducing the number of environment interactions, while mitigating long-term gradient instabilities. Through two challenging parametric navigation problems in unsteady flows, we show that PEARL {\em (i)} effectively exploits differentiable environments to outperform state-of-the-art RL algorithms, {\em (ii)} is sample efficient, thanks to the physics-guided policy learning, {\em (iii)} generalizes across multiple scenarios, which is crucial when dealing with parametric systems, and {\em (iv)} enables scaling RL to high-dimensional state and action spaces, without requiring low-dimensional state representations or multi-agent strategies. 
\end{abstract}
\maketitle


\section{INTRODUCTION} \label{intro}

{\em Reinforcement learning} (RL) is widely gaining traction in solving complex decision-making tasks~\cite{sutton1998reinforcement, bertsekas2019reinforcement, lewis2013reinforcement, sutton1992reinforcement}. Specifically, RL considers an agent in an environment learning optimal decision-making strategies to maximize the expected cumulative reward over time through trial-and-error. The reward is a task-dependent scalar signal -- possibly time-delayed, sparse, and noisy -- evaluating the system response to different agent's actions, which is {\em enough} for discriminating decisions and guiding agent learning~\cite{silver2021}. In the \emph{model-free} RL paradigm, an explicit model of the environment is not available to the agent. Therefore, optimal actuation strategies can only be learned from the experience -- i.e., the data -- obtained through multiple environment interactions. Although broadly applicable in practice, the model-free assumption comes at the cost of severe sample inefficiency, as many simulations -- i.e., interactions with the environment -- are often required to explore (potentially continuous and high-dimensional) action spaces.

In recent years, {\em deep reinforcement learning} (DRL) has emerged as a scalable (feedback) control strategy, capable of coping with continuous states and actions thanks to a neural network-based agent approximation~\cite{li2017deep, arulkumaran2017deep, franccois2018introduction}. DRL has been employed in several applied sciences and engineering fields, such as games~\cite{Silver2016, mnih2013playing, mnih2015human, silver2017masteringchessshogiselfplay, ye2020mastering, shao2019survey, lample2017playing, van2016deep}, simulated and real-world robotics~\cite{lin1992reinforcement, gu2017deep, tang2025, kober2013reinforcement, polydoros2017survey, zhang2015towards, zhao2020sim}, autonomous driving~\cite{Govinda2025}, smart buildings~\cite{Manjavacas2024}, power systems~\cite{glavic2019}, and large language models~\cite{kaufmann2025surveyreinforcementlearninghuman, bai2022traininghelpfulharmlessassistant}. However, neural network training may be data hungry, extremely unstable due to the online nature of the RL setting, and entails an intractable computational cost for complex and high-dimensional tasks.

To enhance data efficiency, {\em model-based reinforcement learning} (MBRL) approaches have been proposed~\cite{sutton1998reinforcement, Luo2024}. Specifically, instead of solely interacting with the (real) environment, the agent investigates and evaluates new decisions using a surrogate model, which is learned from the interaction data and emulates the underlying and unknown dynamics of the environment. Nevertheless, learning an accurate, robust, and computationally efficient proxy of the environment may be extremely challenging or even prohibitive, as in the case of complex systems with distributed actuation, biasing the agent training towards sub-optimal control strategies and degrading its performance when deployed on the real system~\cite{pmlr-v267-barkley25a}. Many strategies have been proposed to overcome the MBRL limitations, thus enhancing robustness, generalization, and interpretability, such as latent world models and planning~\cite{hafner2019learninglatentdynamicsplanning, hafner2020dreamcontrollearningbehaviors, Hafner2025}, short model-based rollouts~\cite{janner2021trustmodelmodelbasedpolicy}, sparse identification of nonlinear dynamics~\cite{Zolman2025, bruntoncontrol2025}, and unsupervised state representation learning~\cite{botteghireview}. 

\begin{figure*}[t]
    \centering
    \includegraphics[width=\linewidth]{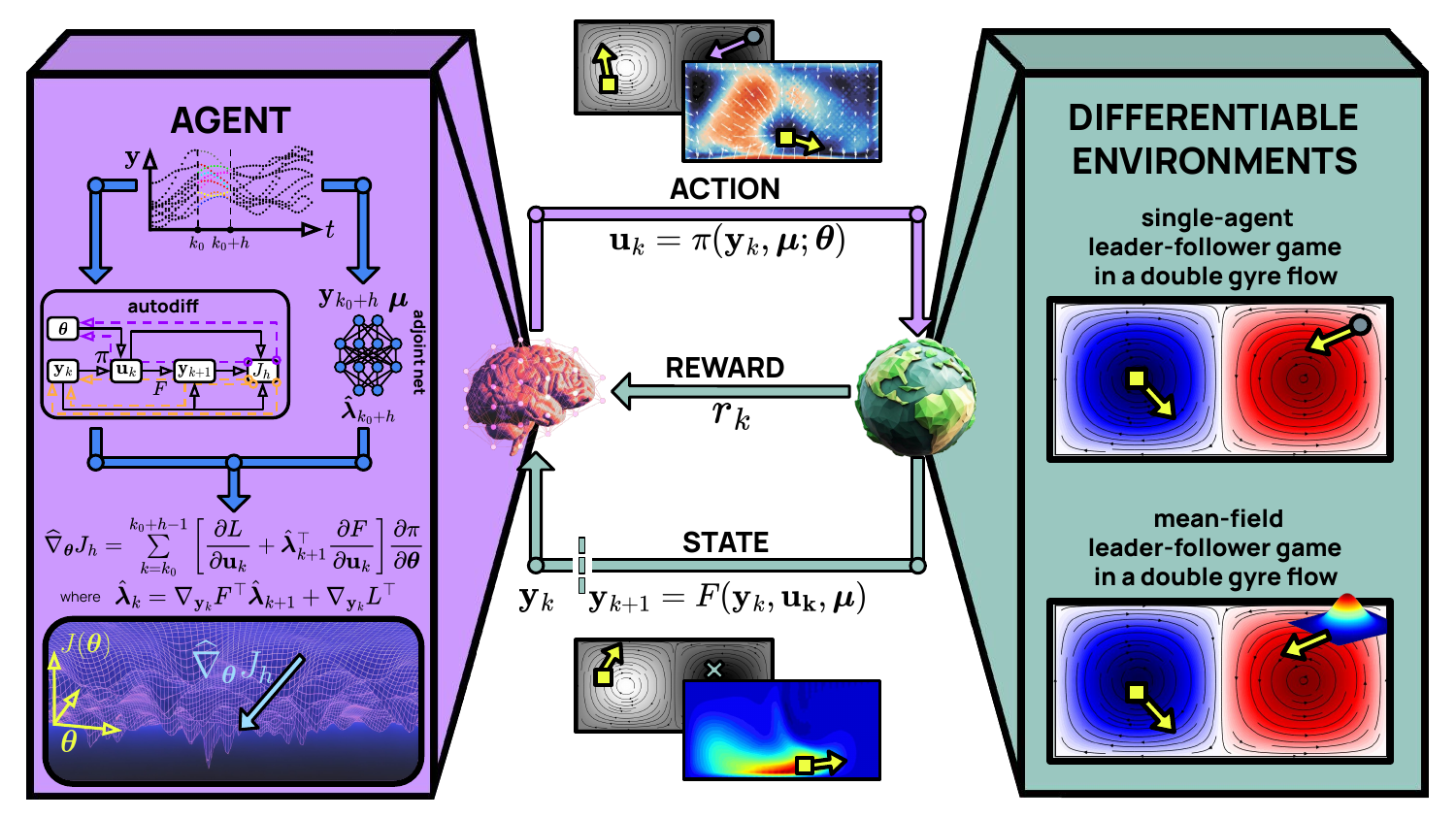}
    \caption{Graphical summary of {\em Physics-EnhAnced Reinforcement Learning} (PEARL). Starting from the current state of the system $\mathbf{y}_k$ and the scenario parameters $\boldsymbol{\mu}$, the agent's policy determines the optimal action to apply to the environment, that is $\mathbf{u}_k = \pi(\mathbf{y}_k, \boldsymbol{\mu};\boldsymbol{\theta})$. As a results, the agent collects the subsequent state $\mathbf{y}_{k+1} = F(\mathbf{y}_k, \mathbf{u}_k,\boldsymbol{\mu})$ and the reward $r_k$, enabling continuous control of the dynamics with the feedback loop. To train the policy, PEARL employs an actor-adjoint algorithm where the gradient of the loss function over the generic short-horizon $k=k_0,\ldots,k_0+h$ is computed via automatic differentiation, thus backpropagating the gradients through the dynamics, and the adjoint equation, whose terminal condition $\hat{\boldsymbol{\lambda}}_{k_0+h}$ is approximated with a neural network to account for long-term dependencies.}
    \label{fig:1}
\end{figure*}

RL has been recently applied to control dynamical systems governed by differential equations, with particular emphasis on fluid dynamics~\cite{bucci2019control, Verma2018, Rabault2019, Ren2021a, fan2020reinforcement, Varela2022, beintema2020controlling, buzzicotti2020optimal, xia2023active}, and plasma confinement~\cite{degrave2022magnetic}. See ~\cite{Vinuesa2022, garnier2021review, rabault2020deep, Vignon2023a, semeraro} for an extensive overview of RL in scientific computing. However, balancing exploration and exploitation -- i.e., trying new actions to gain knowledge about the environment and selecting the best-known actions to maximize immediate reward -- in continuous and high-dimensional action spaces is one of the major challenges in DRL. While standard exploration strategies are effective for low-dimensional control problems, earning high rewards using randomly sampled actions is extremely unlikely when dealing with high-dimensional dynamical systems, thus leading to inaccurate gradient estimates, non-significant agent updates, and premature convergence to local minima. Moreover, determining which action contributed to success or failure, namely the so-called {\em credit assignment problem}, is often prohibitive in this context. To tackle the curse of dimensionality, {\em multi-agent reinforcement learning} (MARL) employs multiple decision-making agents interacting with the environment to achieve certain goals~\cite{bucsoniu2010multi, Busoniu2008, tan1993multi, albrecht2024multi}. Multi-agent systems find applications in a large variety of domains such as, e.g., robotics and swarm systems~\cite{orr2023multi, huttenrauch2019deep}, as well as cooperative and competitive environments~\cite{littman1994markov, tampuu2017multiagent}, and, more recently, multi-agent flow control~\cite{Vignon2023, Vasanth2024, suarez2025active, Peitz2024, hypemarl2025}. However, agents' interactions, communication and coordination, as well as partial or local observability, may compromise convergence and stability~\cite{HernandezLeal2019}.

In contrast to model-free RL, where the environment is treated as a black box, traditional optimal control (OC) theory directly incorporates the governing dynamics of the environment -- which is often explicitly available in this context, at least approximately, through physical laws -- to discover optimal control strategies via physics-constrained optimization problems~\cite{manzoni2021optimal, Bertsekas1995a, kirk2004optimal}. Specifically, a set of Karush-Kuhn-Tucker optimality conditions can be obtained through the Lagrange multipliers' method or the so-called {\em adjoint state method}~\cite{manzoni2021optimal, Bertsekas1995a, kirk2004optimal}, which is closely related to the {\em Pontryagin maximum principle}~\cite{Pontryagin1962}. Note that, in this context, the problem is typically formulated as the minimization of a task-specific loss function, which is equivalent to maximizing the expected cumulative reward, up to a change of sign in the objective. However, the OC paradigm typically focuses on open-loop actuation in deterministic settings, thus preventing generalization to multiple scenarios and allowing only limited robustness to external disturbances. Conversely, physics-based closed-loop control strategies may be designed through the {\em Hamilton-Jacobi-Bellman equation}~\cite{kirk2004optimal, Fleming2006-gv, bardi1997optimal} or {\em model predictive control} (MPC)~\cite{Camacho2004}. However, these solutions become computationally intractable for (possibly parametric) moderate to large-scale dynamical systems. For instance, MPC performs multiple physics-constrained optimizations online, thus introducing latency and violating real-time requirements in challenging high-dimensional settings~\cite{LOHNING20141647, WEINAN2022121}. Similarly to MBRL, in the specific case of low-dimensional actuation, surrogate models may be considered to mitigate complexity, while sacrificing fidelity and undermining control performance. 

Within the scope of RL, physical priors have been mainly utilized to alleviate the curse of dimensionality, ensure safe and efficient explorations, design reward functions, learn low-dimensional state representations, and improve the accuracy of surrogate models in MBRL~\cite{banerjee2023survey}. For instance, Liu and Wang~\cite{Liu2021} exploit governing equations and physical constraints to inform the model learning in MBRL, thus enhancing generalization and performance in the low-data limit. With the advent of differentiable simulators and surrogate models, {\em automatic differentiation} (AD) has been utilized to compute exact cumulative-reward gradients through the dynamics, thus effectively enabling the training of neural network-based open-loop controllers~\cite{Bttcher2022,Holl2020Learning,NEURIPS2020_5a7b238b} and policies~\cite{pilco, pmlr-v139-mora21a, PhysRevLett.134.044001}. For instance, Mokbel et al.~\cite{Mokbel2025} compute the reward function when stabilizing chaotic events in differential fluid simulation via AD. Instead, analytic policy gradients have been exploited by Freeman et al.~\cite{freeman2021braxdifferentiablephysics} and Wiedemann et al.~\cite{wiedemann2023trainingefficientcontrollersanalytic}, while they have been estimated by Eberhard et al.~\cite{pmlr-v283-eberhard25a} to find optimal control actions in model-free open-loop RL. However, when employing AD over long rollouts -- a procedure typically referred to as \emph{backpropagation through time} (BPTT) -- the recursive multiplication of derivatives arising from the chain rule can lead to vanishing or exploding gradients~\cite{Bengio1994}, compromising the optimization. To mitigate gradient instabilities over long horizons, optimization routines over smaller time windows are typically employed, at the cost of introducing a bias. For example, Liu and MacArt~\cite{liu2024} consider multiple policy updates over short horizons for active flow control problems. Moreover, Xu et al.~\cite{xu2022acceleratedpolicylearningparallel} propose the {\em short-horizon actor critic} (SHAC) algorithm for differentiable environments, which mitigates the short-horizon bias by approximating the expected cumulative rewards the agent would accumulate beyond the short horizon with a value function neural network, thus avoiding the learning of myopic policies.

To overcome the limitations of the aforementioned approaches, in this work we bridge the gap between RL and OC by introducing a novel {\em Physics-EnhAnced Reinforcement Learning} (PEARL) algorithm, where we leverage the physical knowledge of the environment to guide the learning process in a RL context, thus enhancing sample efficiency, while enabling optimal and adaptive closed-loop control actions in real-time. In particular, we propose an {\em actor-adjoint algorithm} tailored to the control of differentiable environments. Similarly to SHAC, we exploit AD to compute physics-enhanced gradients over short horizons, thus preventing vanishing or exploding gradients. However, rather than approximating the long-term rewards accumulated beyond the short horizon, we employ a neural network to predict its gradient -- i.e the so-called {\em adjoint} or \emph{value gradient} --, thus accounting for long-term dependencies at the policy gradient level, speeding up optimization and significantly improving sample efficiency. Throughout our numerical experiments, we demonstrate that PEARL outperforms state-of-the-art model-free RL algorithms, such as proximal policy optimization \cite{schulman2017proximal} and twin-delayed deep deterministic policy gradient \cite{fujimoto2018addressing}, and AD-based RL methods, such as BPTT and SHAC. A graphical summary of PEARL is available in Figure~\ref{fig:1}.

The paper is organized as follows. Section~\ref{sec:rlocp} presents a unified overview of RL foundations and classical OC theory for dynamical systems, as well as RL algorithms for differentiable environments. Section~\ref{sec:pearl} proposes a novel RL algorithm tailored to control differentiable environments. Section~\ref{sec:test} assesses the performance of the proposed control strategy in two challenging parametric navigation problems in unsteady flows. Eventually, Section~\ref{sec:conlusions} discusses results and future directions for advancing the proposed methodology toward large-scale and real-world problems.

\section{BRIDGING REINFORCEMENT LEARNING AND OPTIMAL CONTROL}
\label{sec:rlocp}

\subsection{FUNDAMENTALS OF REINFORCEMENT LEARNING}

RL aims to design agents capable of making optimal decisions to maximize the cumulative reward. During training and deployment, agents interact with the underlying environment, which is typically modeled through Markov decision processes (MDPs) to account for stochasticity and uncertainties. Specifically, for a given time instant $t_k$ of the grid $\{t_0, \ldots, t_{N_t}\}$ discretizing the finite horizon $[0,T]$ with final time $T>0$, the agent observes the state of the system $\mathbf{y}_k \in \mathbb{R}^{N_y}$ and takes action $\mathbf{u}_k \in \mathbb{R}^{N_u}$. As a result, the system evolves according to the transition probability $\mathbb{P}(\mathbf{y}_{k+1}|\mathbf{y}_k, \mathbf{u_k}):  \mathbb{R}^{N_y} \times   \mathbb{R}^{N_y} \times  \mathbb{R}^{N_u} \rightarrow [0, 1]$. Additionally, the agent receives the instantaneous scalar reward $r_k$ given by the problem-specific reward function $R(\mathbf{y}_k,\mathbf{u}_k):  \mathbb{R}^{N_y} \times  \mathbb{R}^{N_u} \rightarrow \mathbb{R}$. Importantly, initial conditions $\mathbf{y}_0$, environment dynamics $\mathbb{P}(\mathbf{y}_{k+1}|\mathbf{y}_k, \mathbf{u_k},\boldsymbol{\mu})$, and reward functions $R(\mathbf{y}_k,\mathbf{u}_k,\boldsymbol{\mu})$ may depend on a set of parameters $\boldsymbol{\mu} \in \mathcal{P} \subseteq \mathbb{R}^{N_\mu}$ in the parameter space $\mathcal{P}$, thus requiring different control strategies. Note that, for the sake of simplicity, we consider known and constant scenario parameters in this section, even though time-dependent parameters $\boldsymbol{\mu}(t)$ can also be taken into account, as demonstrated in Section~\ref{sec:test}.

The agent's behavior is defined by the so-called {\em policy} function, which may be either stochastic -- i.e., a distribution over the possible actions, conditioned on the current state -- or deterministic. In the context of DRL, policies are modeled through neural networks with weights and biases $\boldsymbol{\theta} \in \mathbb{R}^{N_\theta}$. In this context, deterministic policies read as follows
\begin{equation*}
    \mathbf{u}_k = \pi(\mathbf{y}_k,\boldsymbol{\mu};\boldsymbol{\theta}) \quad \text{for } \; k = 0,\ldots,N_t.
\end{equation*}
The agent's goal is to find the optimal policy parameters maximizing the total expected cumulative reward $G$ across multiple initial states $\mathbf{y}_0$ and, if present, scenario parameters $\boldsymbol{\mu}$
\begin{equation}
    \underset{\boldsymbol{\theta}}{\max}  G(\boldsymbol{\theta}) = \mathbb{E}_{\pi_{(\cdot;\boldsymbol{\theta})}}\left[\sum_{k=0}^{N_t}\gamma^k r_k\right] = \mathbb{E}_{\substack{\mathbf{y}\sim\rho_0\\\boldsymbol{\mu} \sim \rho_\mu}}[V_{\pi_{(\cdot;\boldsymbol{\theta})}}(\mathbf{y}, \boldsymbol{\mu})],
\label{eq:RLoptim}
\end{equation}
where $\mathbb{E}_{\pi_{(\cdot;\boldsymbol{\theta})}}$ denotes the expected value given the policy $\pi$ driving the system evolution, $\rho_0$ and $\rho_\mu$ are, respectively, probability distributions accounting for different possible initial conditions and system parameters, and $\gamma \in [0, 1]$ indicates the discount factor. Notably, the undiscounted setting, namely $\gamma = 1$, may be only considered for finite-horizon episodic tasks with guaranteed termination. Notice that, thanks to the law of total expectation, it is possible to formulate the optimization in terms of the so-called {\em value function} $V: \mathbb{R}^{N_y}\times\mathcal{P} \rightarrow \mathbb{R}$:
\begin{equation}
    V_\pi(\mathbf{y},\boldsymbol{\mu}) = \mathbb{E}_{\pi}\left[\sum_{k=0}^{N_t} \gamma^k r_k|\mathbf{y}_0=\mathbf{y},\boldsymbol{\mu}\right],
\end{equation}
which measures how good the state $\mathbf{y}$ is in terms of the expected cumulative reward obtained when following the policy $\pi$ starting from the state itself.

RL methods can be classified in three main categories: \emph{(i)} value based, \emph{(ii)} policy based, and \emph{(iii)} actor critic. Value-based approaches aim to estimate the value function $V$ or the \emph{action-value function}
\begin{equation}
    Q_\pi(\mathbf{y},\mathbf{u},\boldsymbol{\mu}) = \mathbb{E}_\pi\left[\sum_{k=0}^{N_t} \gamma^k r_k|\mathbf{y}_0 = \mathbf{y}, \mathbf{u}_0 = \mathbf{u},\boldsymbol{\mu}\right],
\end{equation}
where $Q: \mathbb{R}^{N_y}\times  \mathbb{R}^{N_u}\times \mathcal{P}\rightarrow \mathbb{R}$ is the {\em quality function} assessing the value of state-action pairs under the policy $\pi$ driving decisions. After obtaining the value function, the optimal policy is simply derived by selecting the actions maximizing the value function, leading to the so-called \emph{greedy policy}. Policy gradient methods, instead, maximize the expected cumulative reward directly through gradient ascent algorithms in the policy parameter space. To do so, the {\em policy gradient theorem}~\cite{sutton1999policy} is used to compute the gradient of the performance measure $G$ with respect to the policy parameters $\boldsymbol{\theta}$, that is
\begin{equation}
    \nabla_{\boldsymbol{\theta}} G =
    \mathbb{E}_{\substack{\mathbf{y} \sim \rho_\pi\\\mathbf{u} \sim \pi\\\boldsymbol{\mu}\sim\rho_\mu}}
    \left[
        \nabla_{\boldsymbol{\theta}} \log \pi(\mathbf{u}|\mathbf{y},\boldsymbol{\mu}; \boldsymbol{\theta})
        Q_\pi(\mathbf{y}, \mathbf{u},\boldsymbol{\mu})
    \right],
    \label{eq:pg}
\end{equation}
where $\mathbf{u} \sim \pi(\mathbf{u}|\mathbf{y},\boldsymbol{\mu};\boldsymbol{\theta})$ is the parametrized stochastic policy returning the action probability $\mathbb{P}(\mathbf{u}|\mathbf{y},\boldsymbol{\mu})$. Whenever a deterministic policy is employed, the gradient expression is given, instead, by the {\em deterministic policy gradient theorem}~\cite{silver2014deterministic}
\begin{equation}
    \nabla_{\boldsymbol{\theta}} G =
    \mathbb{E}_{\substack{\mathbf{y} \sim \rho_\pi\\\boldsymbol{\mu}\sim\rho_\mu}}
    \left[
        \left. \nabla_{\mathbf{u}} Q_\pi(\mathbf{y}, \mathbf{u},\boldsymbol{\mu}) \right|_{\mathbf{u}=\pi(\mathbf{y},\boldsymbol{\mu};\boldsymbol{\theta})} \nabla_{\boldsymbol{\theta}} \pi(\mathbf{y},\boldsymbol{\mu};\boldsymbol{\theta}) 
    \right].
    \label{eq:dpg}
\end{equation}
In practice, the gradients in Equations~\eqref{eq:pg} and \eqref{eq:dpg} are approximated through empirical means based on the trajectories sampled during training. For instance, the REINFORCE algorithm~\cite{williams1992simple} exploits an unbiased policy gradient approximation via Monte Carlo returns, at the cost of high variance, thus making learning unstable and data inefficient. Indeed, as shown in Appendix~\ref{sec:appendix1}, the rate of convergence of gradient ascent algorithms with inexact gradients is $O\left(\frac{\sigma}{\sqrt{N}}\right)$, where $\sigma^2$ is the variance of the gradient estimator and $N$ is the number of iterations, resulting in a sample complexity of $N = O\left(\frac{\sigma^2}{\varepsilon^2}\right)$ to reach convergence up to a prescribed tolerance $\varepsilon > 0$. Different strategies have been proposed to reduce the variance of the gradient estimator, while introducing bias due to the {\em bias-variance trade-off}~\cite{variancered}, such as control variates, advantage function approximation~\cite{schulman2018highdimensionalcontinuouscontrolusing}, and actor-critic strategies~\cite{actorcritic}. Actor-critic algorithms such as, e.g., {\em proximal policy optimization} (PPO)~\cite{schulman2017proximal} and {\em twin-delayed deep deterministic policy gradient} (TD3)~\cite{fujimoto2018addressing} aim at jointly learning a policy -- i.e., the actor -- and a critic -- i.e., the value function -- with the goal of improving upon value- and policy-based methods. However, $\sigma^2$ generally scales with the amount of policy parameter $N_\theta$~\cite{Nesterov2015,Ghadimi2013}, as well as with the final time $T$ and the problem dimensions $N_y$, $N_u$~\cite{agarwal2020theorypolicygradientmethods, wu2018variancereductionpolicygradient, Peters2008, pmlr-v162-suh22b}, thus making model-free RL extremely inefficient, especially for high-dimensional problems over long horizons.

\subsection{OPTIMAL CONTROL OF PARAMETRIC DYNAMICAL SYSTEMS}

In this work, we take into account deterministic environments, such as those described by means of nonlinear, high-dimensional and parametric ordinary or partial differential equations (ODEs/PDEs) in the form
\begin{equation*}
\dot{\mathbf{y}} = f(\mathbf{y},\mathbf{u},\boldsymbol{\mu}).
\end{equation*}
Rather than relying on transition probabilities, the temporal evolution of the state $\mathbf{y}(t) \in \mathbb{R}^{N_y}$ over $t \in [0,T]$ under control actions $\mathbf{u}(t) \in \mathbb{R}^{N_u}$ is thus deterministic and governed by the vector field $f:\mathbb{R}^{N_y}\times\mathbb{R}^{N_u}\times\mathcal{P} \to \mathbb{R}^{N_y}$ modeling its time derivative $\dot{\mathbf{y}}$. Note that the presented framework may be easily extended to stochastic or partially observable systems. Note also that, in the case of spatio-temporal systems, spatial discretization of the domain $\Omega$ through, e.g., finite element, finite volume or spectral methods~\cite{kutz2013data, courant2008methods} is required to deal with finite-dimensional variables. Moreover, similarly to MDPs, we discretize the dynamics over time, leading to
\begin{equation}
\mathbf{y}_{k+1} = F(\mathbf{y}_k,\mathbf{u}_k,\boldsymbol{\mu}),
\label{eq:PDEdiscrete}
\end{equation}
where $\mathbf{y}_k = \mathbf{y}(t_k)$ and $\mathbf{u}_k = \mathbf{u}(t_k)$ for $k=0,\ldots,N_t$, while $F:\mathbb{R}^{N_y}\times\mathbb{R}^{N_u}\times\mathcal{P} \to \mathbb{R}^{N_y}$ is the discrete-time transition function.

When dealing with deterministic environments with evolution described by Equation~\eqref{eq:PDEdiscrete}, the optimization problem in Equation~\eqref{eq:RLoptim} is analogous to the following constrained optimization~\cite{semeraro, Brunton2022} -- typically referred to as {\em optimal control problem} (OCP)~\cite{manzoni2021optimal, Bertsekas1995a, kirk2004optimal}
\begin{equation}
\begin{aligned}
&\underset{\boldsymbol{\theta}}{\min} J(\boldsymbol{\theta}) = \sum_{k=0}^{N_t-1} L(\mathbf{y}_k, \pi(\mathbf{y}_k,\boldsymbol{\mu};\boldsymbol{\theta}),\boldsymbol{\mu}) + \phi(\mathbf{y}_{N_t},\boldsymbol{\mu}),
\\
&\text{ s.t.} \; \mathbf{y}_{k+1} = F(\mathbf{y}_k, \pi(\mathbf{y}_k,\boldsymbol{\mu};\boldsymbol{\theta}), \boldsymbol{\mu}) \quad  \text{for } \; k = 0,\ldots,N_t-1,
\end{aligned}
\label{eq:ocpdiscrete}
\end{equation}
where the so-called {\em cost} or {\em loss function} $J$, with $L:\mathbb{R}^{N_y} \times \mathbb{R}^{N_u} \times \mathcal{P} \to \mathbb{R}$ the running cost and $\phi:\mathbb{R}^{N_y} \times \mathcal{P} \to \mathbb{R}$ the final payoff, is negative of the performance measure $G$ in deterministic and undiscounted settings. Notice that, while Equation~\eqref{eq:ocpdiscrete} searches for optimal policy parameters across multiple scenarios and initial conditions, conventional OCPs directly optimize control actions in fixed configurations. 

When the environment dynamics is available, the gradient of the loss function in Equation~\eqref{eq:ocpdiscrete} can be computed exactly through the adjoint state method~\cite{manzoni2021optimal, Bertsekas1995a, kirk2004optimal} or, equivalently, backpropagation through time (BPTT)~\cite{werbos2002backpropagation} and reverse-mode automatic differentiation (AD)~\cite{10.1007/3-540-28438-9_2, baydin2018automatic, margossian2019review}, yielding
\begin{equation}
\nabla_{\boldsymbol{\theta}} J = \sum_{k=0}^{N_t-1} \left[\dfrac{\partial L}{\partial \mathbf{u}_k} + \boldsymbol{\lambda}_{k+1}^{\top} \dfrac{\partial F}{\partial \mathbf{u}_k}\right] \dfrac{\partial \pi}{\partial \boldsymbol{\theta}}
\label{eq:grad2}
\end{equation}
where $\boldsymbol{\lambda}_k(\boldsymbol{\mu})\in \mathbb{R}^{N_y}$ for $k=0,\ldots,N_t$ are the closed-loop {\em adjoint} variables satisfying the (linear) adjoint equation running backward in time, that is
\begin{equation}
\boldsymbol{\lambda}_{k} = \left( \dfrac{\partial F}{\partial \mathbf{y}_k} + \dfrac{\partial F}{\partial \mathbf{u}_k}\dfrac{\partial \pi}{\partial \mathbf{y}_k}\right)^\top \boldsymbol{\lambda}_{k+1} + \left(\dfrac{\partial L}{\partial \mathbf{y}_k} + \dfrac{\partial L}{\partial \mathbf{u}_k}\dfrac{\partial \pi}{\partial \mathbf{y}_k}\right)^\top
\label{eq:adj2}
\end{equation}
with terminal condition $\boldsymbol{\lambda}_{N_t} = \frac{\partial \phi}{\partial \mathbf{y}_{N_t}}$. See Appendix~\ref{sec:appendix2} for the derivation of Equation~\eqref{eq:grad2} through the adjoint state method and Appendix~\ref{sec:appendix3} for the equivalence with respect to reverse-mode AD. Importantly, the computational cost of adjoint problems is independent of the number of policy parameters $N_\theta$, in contrast to direct or forward-mode approaches. Moreover, in contrast to inexact gradient methods, first-order optimization with exact gradients entails a convergence rate equal to $O\left( \frac{1}{N} \right)$, yielding a sample complexity equal to $O\left( \frac{1}{\varepsilon} \right)$~\cite{nesterov2013introductory, Ghadimi2013, bertsekas2016nonlinear}, thus allowing us to cope with continuous and high-dimensional problems, as demonstrated in Section~\ref{sec:test}. 

\subsection{ADJOINT VARIABLE AS A SENSITIVITY MEASURE}

This section delves into the role of the adjoint variable as a sensitivity measure, both in continuous and discrete-time, to enhance intuition and interpretability regarding physics-enhanced optimization, as well as to further bridge the gap between RL and traditional OC theory. For the sake of generality, we first consider the continuous-time extension of the OCP in Equation~\eqref{eq:ocpdiscrete}, that is
\begin{equation}
\begin{aligned}
&\underset{\boldsymbol{\theta}}{\min} J(\boldsymbol{\theta}) = \int_0^T L(\mathbf{y}(t), \pi(\mathbf{y}(t),\boldsymbol{\mu};\boldsymbol{\theta}),\boldsymbol{\mu}) dt + \phi(\mathbf{y}(T),\boldsymbol{\mu}),
\\
&\text{ s.t. } \; \dot{\mathbf{y}} = f(\mathbf{y}, \pi(\mathbf{y},\boldsymbol{\mu};\boldsymbol{\theta}), \boldsymbol{\mu}).
\end{aligned}
\label{eq:ocpcont}
\end{equation}
As demonstrated in Appendix~\ref{sec:appendix2}, it is possible to compute the exact gradient of the loss function $J$ with respect to the policy parameters $\boldsymbol{\theta}$ through the adjoint state method, yielding 
\begin{equation}
\nabla_{\boldsymbol{\theta}} J =\int_0^T \left(\dfrac{\partial L}{\partial \mathbf{u}} + \boldsymbol{\lambda}^{\top} \dfrac{\partial f}{\partial \mathbf{u}}\right)\dfrac{\partial \pi}{\partial \boldsymbol{\theta}} dt,
\label{eq:grad1}
\end{equation}
where $\boldsymbol{\lambda}(t,\boldsymbol{\mu}) \in \mathbb{R}^{N_y}$ is the closed-loop adjoint variable satisfying the (linear) adjoint ODE
\begin{equation}
\dot{\boldsymbol{\lambda}} = -\left( \dfrac{\partial f}{\partial \mathbf{y}} + \dfrac{\partial f}{\partial \mathbf{u}}\dfrac{\partial \pi}{\partial \mathbf{y}}\right)^\top\boldsymbol{\lambda} - \left( \dfrac{\partial L}{\partial \mathbf{y}} + \dfrac{\partial L}{\partial \mathbf{u}}\dfrac{\partial \pi}{\partial \mathbf{y}}\right)^\top
\label{eq:adj1}
\end{equation}
with terminal condition $\boldsymbol{\lambda}(T,\boldsymbol{\mu}) = \frac{\partial \phi}{\partial \mathbf{y}}^{\top}(\mathbf{y}(T),\boldsymbol{\mu})$. As demonstrated in Appendix~\ref{sec:appendix4}, the adjoint variable $\boldsymbol{\lambda}(t,\boldsymbol{\mu})$ corresponds to the sensitivity of the value function to the state $\mathbf{y}$, that is
\begin{equation}
\boldsymbol{\lambda}(t,\boldsymbol{\mu}) = \nabla_\mathbf{y}V_\pi(\mathbf{y},t,\boldsymbol{\mu})^\top  \; \text{for } \, t \in [0,T],
\label{eq:valuegrad}
\end{equation}
where, in this context, the continuous-time value function $V_\pi(\mathbf{y},t,\boldsymbol{\mu})$ under a generic fixed policy $\pi$ reads as follows
\begin{equation}
V_\pi(\mathbf{y},t,\boldsymbol{\mu}) = \int_t^TL(\mathbf{y}(\tau),\pi(\mathbf{y}(\tau),\boldsymbol{\mu};\boldsymbol{\theta}),\boldsymbol{\mu})d\tau + \phi(\mathbf{y}(T),\boldsymbol{\mu}).
\label{eq:valueocp}
\end{equation}
Significantly, the optimal value function $V(\mathbf{y},t,\boldsymbol{\mu})$ associated with the optimal policy is given by the {\em Hamilton-Jacobi-Bellman equation}~\cite{kirk2004optimal, Fleming2006-gv, bardi1997optimal}
\[
-\dfrac{\partial V}{\partial t} = \underset{\mathbf{u}(t)}{\min}[L(\mathbf{y}, \mathbf{u},\boldsymbol{\mu}) + \nabla_\mathbf{y} V f(\mathbf{y}, \mathbf{u},\boldsymbol{\mu}))]
\]
to be solved backward in time, with terminal condition $V(\mathbf{y},T,\boldsymbol{\mu}) = \phi(\mathbf{y}(T),\boldsymbol{\mu})$.

After time discretization, the OCP turns into Equation~\eqref{eq:ocpdiscrete}, with exact policy gradient in Equation~\eqref{eq:grad2} and adjoint variables $\boldsymbol{\lambda}_k(\boldsymbol{\mu})=\boldsymbol{\lambda}(t_k, \boldsymbol{\mu})$ for $k=0,\ldots,N_t$ solving Equation~\eqref{eq:adj2}. In this setting, as shown in Appendix~\ref{sec:appendix4}, the adjoint variables are equal to the sensitivity of the value function or the loss $J$ with respect to the state $\mathbf{y}_k$, that is
\begin{equation}
\boldsymbol{\lambda}_k(\boldsymbol{\mu}) = \nabla_{\mathbf{y}_k} V_\pi (\mathbf{y}_k, \boldsymbol{\mu})^\top = \nabla_{\mathbf{y}_k} J^\top \; \text{for } \, k =0,\ldots,N_t,
\label{eq:valuegraddiscrete}
\end{equation}
where the value function at the discrete level $V_\pi (\mathbf{y}_k, \boldsymbol{\mu})$ for $k=0,\ldots,N_t$ is defined as 
\begin{equation}
V_\pi (\mathbf{y}_k, \boldsymbol{\mu}) = \sum_{\kappa = k}^{N_t-1} L(\mathbf{y}_\kappa, \pi(\mathbf{y}_\kappa, \boldsymbol{\mu}; \boldsymbol{\theta}),\boldsymbol{\mu}) + \phi(\mathbf{y}_{N_t},\boldsymbol{\mu}).
\label{eq:valueocpdiscrete}
\end{equation}
Notably, it is often convenient to write the value function in recursive form, thus obtaining the so-called {\em Bellman equation}~\cite{Bellman1957} in deterministic and undiscounted settings
\[
V_\pi (\mathbf{y}_k, \boldsymbol{\mu}) =  L(\mathbf{y}_k, \pi(\mathbf{y}_k, \boldsymbol{\mu}; \boldsymbol{\theta}),\boldsymbol{\mu}) + V_\pi(\mathbf{y}_{k+1},\boldsymbol{\mu}),
\]
which turns into the {\em optimal Bellman equation}~\cite{Bellman1957} for the optimal value function $V (\mathbf{y}_k, \boldsymbol{\mu})$ associated with the optimal policy, that is 
\[
V (\mathbf{y}_k, \boldsymbol{\mu}) = \underset{\pi}{\min}[  L(\mathbf{y}_\kappa, \pi(\mathbf{y}_\kappa, \boldsymbol{\mu}; \boldsymbol{\theta}),\boldsymbol{\mu}) + V_\pi(\mathbf{y}_{k+1},\boldsymbol{\mu})].
\]

\subsection{SHORT-HORIZON POLICY OPTIMIZATION}

When exploiting BPTT to learn optimal policy parameters over long rollouts ($N_t \gg 1$), the recursive multiplication between derivatives can lead to vanishing or exploding gradients~\cite{Bengio1994}, severely compromising the optimization. To overcome this limitation, it is possible to truncate the loss computation, namely the cost/reward accumulation, over shorter horizons of $h \ll N_t$ steps, that is
\[
J_h(\boldsymbol{\theta}) = \sum_{k=k_0}^{k_0 + h - 1} L(\mathbf{y}_k, \pi(\mathbf{y}_k,\boldsymbol{\mu};\boldsymbol{\theta}),\boldsymbol{\mu}) + \phi(\mathbf{y}_{k_0 + h},\boldsymbol{\mu}),
\]
where $k_0 = nh$ is the starting point of the $n^{th}$ short horizon for $n=0,1,\ldots, N_h$, and to optimize $J_h$ in every short horizon. Specifically, one can consider gradient descent with descent direction
\[
\nabla_{\boldsymbol{\theta}} J_h = \sum_{k=k_0}^{k_0 + h - 1} \left[ \dfrac{\partial L}{\partial \mathbf{u}_k} + \boldsymbol{\lambda}^{\top}_{k+1} \dfrac{\partial F}{\partial \mathbf{u}_k} \right]\dfrac{\partial \pi}{\partial \boldsymbol{\theta}},
\]
where the adjoint variables are computed through the adjoint recursion introduced in Equation~\eqref{eq:adj2}, starting from the terminal condition $\boldsymbol{\lambda}_{k_0 + h} = \frac{\partial \phi}{\partial \mathbf{y}_{k_0+h}}$. Importantly, the adjoint variables here depend exclusively on the system dynamics within the short horizon, with no contribution from future information. Therefore, this strategy -- commonly referred to as {\em truncated BPTT} -- is capable of preventing gradient instabilities over long rollouts at the expense of introducing biased gradients and degraded long-term credit assignment.

To mitigate the limitation of truncated BPTT, Xu et al.~\cite{xu2022acceleratedpolicylearningparallel} propose the {\em short-horizon actor critic} (SHAC) algorithm for differentiable environments where a critic network $\psi: \mathbb{R}^{N_y} \times \mathcal{P} \to \mathbb{R}$ with weights and biases $\bar{\boldsymbol{\eta}} \in \mathbb{R}^{N_\eta}$ corrects the value of the short-horizon terminal state $\mathbf{y}_{k_0 + h}$ in order to account for future dependencies, that is
\[
\hat{V}(\mathbf{y}_{k_0 + h},\boldsymbol{\mu}) =  \phi(\mathbf{y}_{k_0 + h},\boldsymbol{\mu})  +  \psi(\mathbf{y}_{k_0+h}, \boldsymbol{\mu};\bar{\boldsymbol{\eta}}),
\]
thus yielding the short-horizon loss
\[
\hat{J}_h(\boldsymbol{\theta}) = \sum_{k=k_0}^{k_0 + h - 1} L(\mathbf{y}_k, \pi(\mathbf{y}_k,\boldsymbol{\mu};\boldsymbol{\theta}),\boldsymbol{\mu}) + \hat{V}(\mathbf{y}_{k_0 + h},\boldsymbol{\mu}).
\]
Note that, when considering the final payoff, the critic network prediction is added to $\phi$ instead of estimating the whole value function. To improve training stability and mitigate oscillations due to rapidly changing value estimates, SHAC considers two copies of the critic networks: the online network with parameters $\boldsymbol{\eta}$, which is trained by minimizing the mean squared error 
\begin{equation}
    J_\psi(\boldsymbol{\eta}) = \dfrac{1}{h} \sum_{k=k_0}^{k_o + h - 1} \left\lVert \bar{V}_k - \hat{V}(\mathbf{y}_k,\boldsymbol{\mu};\boldsymbol{\eta}) \right\rVert^2,
    \label{eq:value_loss}
\end{equation}
where $\lVert \cdot \rVert$ denotes the Euclidean norm, the target values $\bar{V}_k$ are computed via temporal-difference learning~\cite{sutton1998reinforcement}, and a target network with parameters $\bar{\boldsymbol{\eta}}$, which is employed to correct the loss function $J_h$ and to compute the target values $\bar{V}_k$~\cite{mnih2015human}. The target network is not update minimizing the mean square error in Equation~\eqref{eq:value_loss} but, instead, softly updated using the online network parameters $\boldsymbol{\eta}$. Despite providing an efficient and stable learning paradigm, the AD computation of the gradient $\nabla_{\boldsymbol{\theta}} \hat{J}_h$ relies on the value gradient $\nabla_{\mathbf{y}_{k_0 + h}}\hat{V}(\mathbf{y}_{k_0 + h},\boldsymbol{\mu})$, which is derived by backpropagating through the critic network $\psi$. However, the value gradient (i.e., the adjoint) at the short-horizon terminal state $\mathbf{y}_{k_0+h}$ can still be inaccurate even when the value itself is well approximated. The value gradient approximation can therefore be improved by leveraging the underlying physics via the adjoint equation in Equation~\eqref{eq:adj2}, as proposed in the following section.

\section{PHYSICS-ENHANCED REINFORCEMENT LEARNING}
\label{sec:pearl}

In this section, we introduce the {\em Physics-EnhAnced Reinforcement Learning} (PEARL) algorithm tailored to control high-dimensional, distributed and parametric dynamical systems. To mitigate vanishing and exploding gradients over long rollouts, PEARL updates the policy parameters online after every short horizon of length $H \ll T$, i.e., every $h \ll N_t$ time steps, while employing AD to compute the gradient descent directions that drive policy learning. Note that differentiable simulators are needed to backpropagate through the dynamical systems and compute derivatives. Similarly to SHAC, we take into account a neural network-based correction to mitigate the bias introduced with the short-horizon optimization. However, instead of approximating the value of the short-horizon terminal state $\hat{V}(\mathbf{y}_{k_0 + h},\boldsymbol{\mu})$, we directly estimate the adjoint -- i.e., the value gradient -- $\hat{\boldsymbol{\lambda}}_{k_0+h}$ that enters in the descent direction in order to account for long-term dependencies not observed yet, resulting in a value gradient learning strategy~\cite{fairbank2008reinforcementlearningvaluegradients} or an {\em actor-adjoint method}. This paradigm shift is beneficial to achieve physics-informed value gradient estimates and, thus, more effective policy gradients. Indeed, rather than taking into account a temporal-difference (TD) scheme to compute target values and train the critic network, we exploit the adjoint equation in Equation~\eqref{eq:adj2} to generate target adjoint variables and train the so-called {\em adjoint network}, better accounting for how state perturbations propagate through the underlying physical model.

\subsection{ACTOR-ADJOINT METHOD}

In this section, we detail the actor-adjoint method employed by PEARL to learn effective policies for steering dynamical systems, while drastically improving sample efficiency, robustness,  and tackling the curse of dimensionality when compared to several state-of-the-are algorithms, such as PPO, TD3, BPTT, and SHAC. The policy parameters $\boldsymbol{\theta}$ are updated online, while interacting with the environment, after every short horizon of $h \ll N_t$ time steps, towards the gradient descent direction
\begin{equation}
\widehat{\nabla}_{\boldsymbol{\theta}} J_h = \sum_{k=k_0}^{k_0 + h - 1} \left[ \dfrac{\partial L}{\partial \mathbf{u}_k} + \hat{\boldsymbol{\lambda}}^{\top}_{k+1} \dfrac{\partial F}{\partial \mathbf{u}_k} \right]\dfrac{\partial \pi}{\partial \boldsymbol{\theta}},
\label{eq:gradSH}
\end{equation}
which approximates the exact policy gradient in Equation~\eqref{eq:grad2} over short horizons to prevent instabilities. The adjoint approximations $\hat{\boldsymbol{\lambda}}_{k}$ for $k=k_0+1,\ldots,k_0+h-1$ in Equation~\eqref{eq:gradSH} are given in cascade by the adjoint recursion in Equation~\eqref{eq:adj2}, with terminal condition $\hat{\boldsymbol{\lambda}}_{k_0+h}$ approximated through the so-called adjoint network $\varphi:\mathbb{R}^{N_y} \times \mathcal{P} \to \mathbb{R}^{N_y}$ having weights and biases $\bar{\boldsymbol{\vartheta}} \in \mathbb{R}^{N_\vartheta}$, that is
\begin{equation}
\hat{\boldsymbol{\lambda}}_{k_0+h} = \dfrac{\partial \phi}{\partial \mathbf{y}_{k_0+h}} + \varphi(\mathbf{y}_{k_0+h}, \boldsymbol{\mu};\bar{\boldsymbol{\vartheta}}).
\label{eq:adjnet}
\end{equation}
We recall that, as discussed in Section~\ref{sec:rlocp}, the adjoint variables quantify the sensitivity of long-term losses (or rewards) to the current state. Therefore, when optimizing over short horizon, the adjoint variables are not available online due to their backward-in-time nature, and have to be approximated.

Similarly to SHAC, to improve training stability and reduce oscillations caused by rapidly changing targets, we consider two copies of the adjoint network: an online network with parameters $\boldsymbol{\vartheta}$ and a target one with parameters $\bar{\boldsymbol{\vartheta}}$. Specifically, the former is updated through gradient descent after each short horizon, minimizing the mean squared error
\begin{equation*}
J_\varphi(\boldsymbol{\vartheta}) = \dfrac{1}{h}\sum_{k=k_0}^{k_0+h-1} \left\lVert \bar{\boldsymbol{\lambda}}_{k} - \varphi(\mathbf{y}_{k}, \boldsymbol{\mu};\boldsymbol{\vartheta}) \right\rVert^2,
\end{equation*} 
, whereas the latter is employed to predict adjoint variables through Equation~\eqref{eq:adjnet} and to compute target adjoint variables $\bar{\boldsymbol{\lambda}}_{k}$ for $k=k_0,\ldots,k_0+h-1$. The target adjoint variables are computed through a TD scheme based upon the adjoint equation in Equation~\eqref{eq:adj2}, thus providing sharper physics-enhanced gradients tied to the dynamics and more accurate long-term sensitivities. In particular, we consider the TD-$\lambda$ scheme for $k=k_0,\ldots,k_0+h-1$:
\begin{equation}
\begin{aligned}
\bar{\boldsymbol{\lambda}}_{k} = \left( \dfrac{\partial F}{\partial \mathbf{y}_k} + \dfrac{\partial F}{\partial \mathbf{u}_k}\dfrac{\partial \pi}{\partial \mathbf{y}_k}\right)^\top  & \Big( \dfrac{\partial L}{\partial \mathbf{y}_{k+1}} + \lambda \bar{\boldsymbol{\lambda}}_{k+1} + \\&  + (1-\lambda) \varphi(\mathbf{y}_{k+1}, \boldsymbol{\mu};\bar{\boldsymbol{\vartheta}}) \Big),
\end{aligned}
\label{eq:tdlambda}
\end{equation}
where $\frac{\partial L}{\partial \mathbf{y}_{k_0+h}} = \frac{\partial \phi}{\partial \mathbf{y}_{k_0+h}}$ for the sake of compactness. Differently from SHAC, which considers TD schemes for the value function, Equation~\eqref{eq:tdlambda} allows us to provide physics-informed target value gradients when training the adjoint network $\varphi$. In particular, Equation~\eqref{eq:tdlambda} combines the current target adjoint variables and the corresponding adjoint net estimates in the right hand side through a parameter $\lambda \in [0,1]$, enabling credit assignment for longer temporal dependencies. Note that the differences between Equation~\eqref{eq:tdlambda} and the adjoint recursion in Equation~\eqref{eq:adj2} are due to the role of the adjoint net in Equation~\eqref{eq:adjnet}, which does not predict the entire adjoint variable but corrects the loss gradient available within the short horizon. Note also that, when considering a discount $\gamma \in [0, 1)$, the target adjoint variables are rescaled by a factor $\gamma^{k-k_0}$ to remove the discounting contribution from the adjoint dynamics, thus facilitating the learning of the underlying value gradient by the adjoint network. The target adjoint network is then softly updated after each short horizon via the rule
\[
\bar{\boldsymbol{\vartheta}} \leftarrow \alpha \bar{\boldsymbol{\vartheta}} + (1 - \alpha)\boldsymbol{\vartheta} 
\]
with $\alpha \in [0,1)$.

\section{NUMERICAL RESULTS}
\label{sec:test}

In this section, we assess the performance of PEARL in two parametric optimal navigation problems in unsteady flows, which are referred to as \emph{leader-follower game} and \emph{mean-field leader follower game}. Specifically, in Section~\ref{subsec:lfg} we aim at steering an agent -- i.e., the follower -- in a double gyre flow to track a target -- i.e., the leader -- passively transported by the underlying flow field. Note that the initial position of the leader and the follower may vary within the domain, therefore requiring a parametric policy capable of adapting to multiple scenarios. Moreover, Section~\ref{subsec:mean-field_lfg} considers a high-dimensional variation of the leader-follower game, where we apply a distributed control action to steer a Gaussian density to follow a leader moving in the double gyre flow.

In our numerical experiments, to deal with parametric control problems, a neural network architecture composed of two different branches to embed the state and the scenario parameters, which are modeled via feed-forward neural networks with $2$ layers and a problem-specific number of neurons. After encoding the state and the parameters, the corresponding embeddings are concatenated and further processed through a third $2$-layers neural network. For fairness of comparison, the same architecture is used for all the different algorithms. Moreover, as far as the hyperparameters are concerned, we employ a discount factor $\gamma = 0.99$, TD scheme with $\lambda = 0.95$, and target network update with $\alpha = 0.995$ both for SHAC and PEARL.

\subsection{LEADER-FOLLOWER GAME}
\label{subsec:lfg}

\begin{figure*}[t]
    \centering \subfloat{\includegraphics[width=0.5\linewidth]{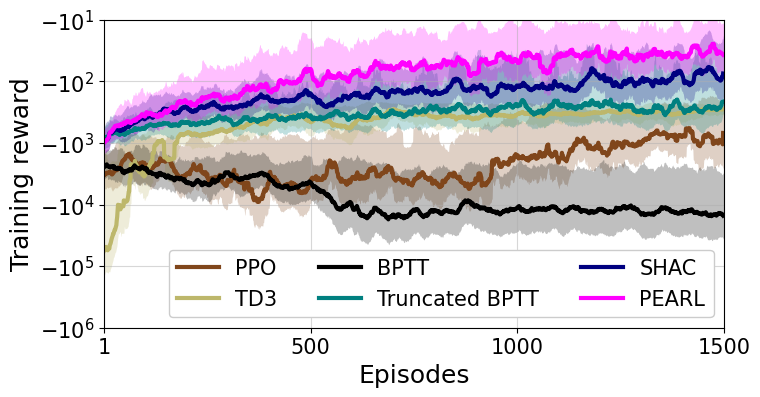}} \subfloat{\includegraphics[width=0.5\linewidth]{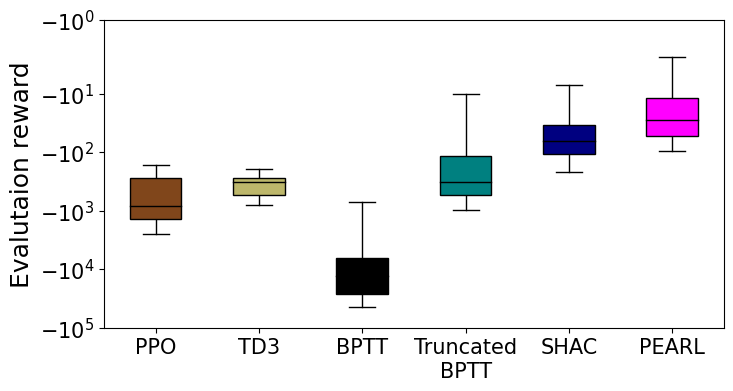}} \\
    
    \subfloat{\includegraphics[width=0.5\linewidth]{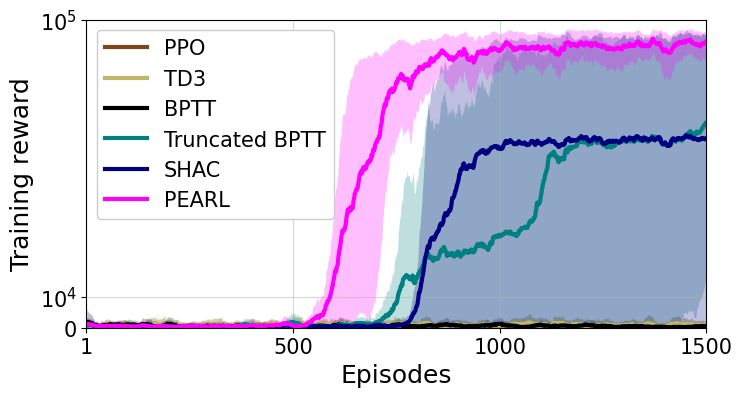}} \subfloat{\includegraphics[width=0.5\linewidth]{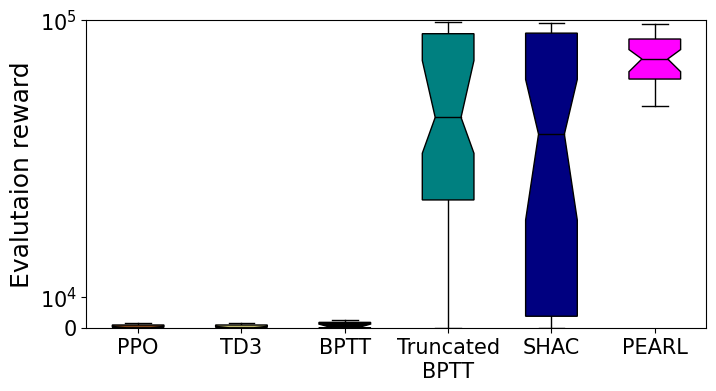}}
    \caption{{\em Leader-follower game}. Training and evaluation rewards obtained by the different competing agents in the leader-follower game with dense and sparse rewards.}
    \label{fig:singleagent_rew}
\end{figure*}

\begin{figure*}[t]
    \centering
    \begin{sideways}
    \makebox[0pt][l]{\hspace{-4.1cm}
    \begin{minipage}{4cm}{\large \bfseries 0 seconds} \end{minipage}}
    \end{sideways}\subfloat[\large \bfseries Test 1]{
        \includegraphics[width=0.33\linewidth, height=0.2\linewidth]{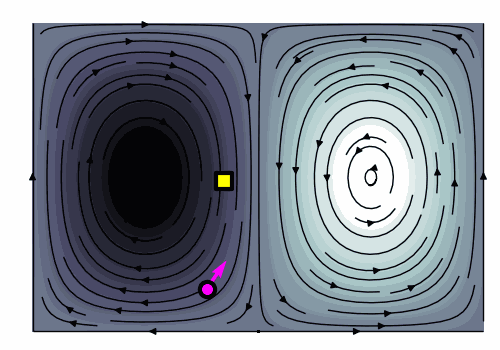}
      
    }
    \subfloat[\large \bfseries Test 2]{
        \includegraphics[width=0.33\linewidth, height=0.2\linewidth]{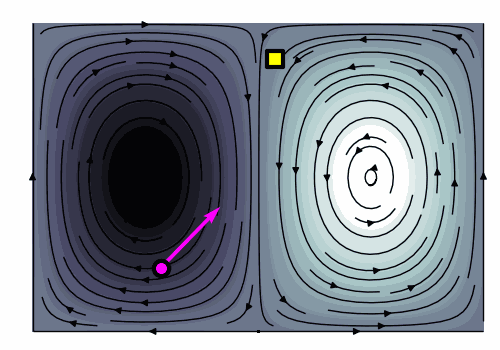}
      
    }
    \subfloat[\large \bfseries Test 3]{
        \includegraphics[width=0.33\linewidth, height=0.2\linewidth]{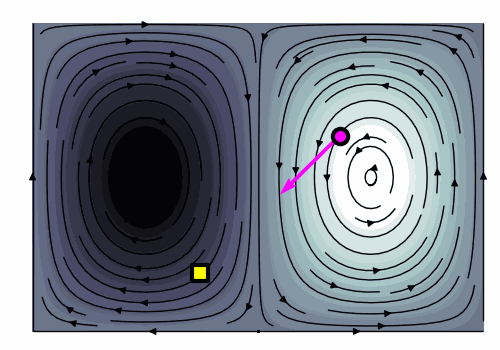}
      
    }

    \vspace{-0.5cm} 
    \begin{sideways}
    \makebox[0pt][l]{\hspace{-4.1cm}
    \begin{minipage}{4cm}{\large \bfseries 5 seconds} \end{minipage}}
    \end{sideways}\subfloat{
        \includegraphics[width=0.33\linewidth, height=0.2\linewidth]{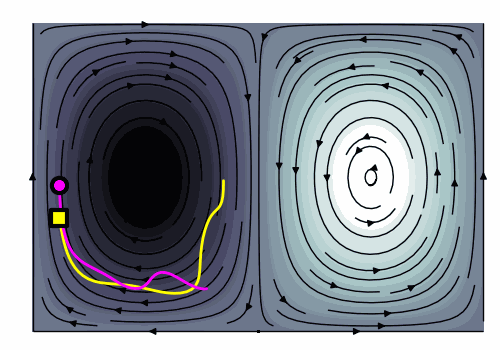}
      
    }
    \subfloat{
        \includegraphics[width=0.33\linewidth, height=0.2\linewidth]{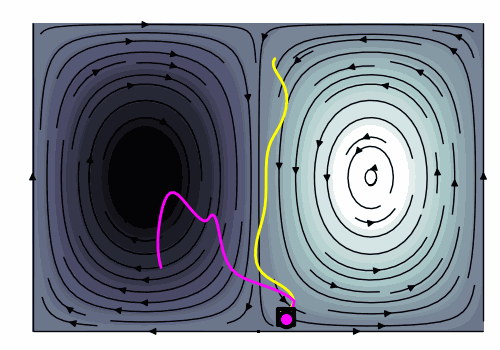}
      
    }
    \subfloat{
        \includegraphics[width=0.33\linewidth, height=0.2\linewidth]{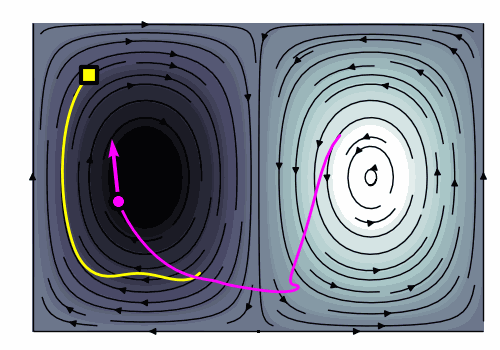}
      
    }

    \vspace{-0.5cm} 
    \begin{sideways}
    \makebox[0pt][l]{\hspace{-4.07cm}
    \begin{minipage}{4cm}{\large \bfseries 10 seconds} \end{minipage}}
    \end{sideways}\subfloat{
        \includegraphics[width=0.33\linewidth, height=0.2\linewidth]{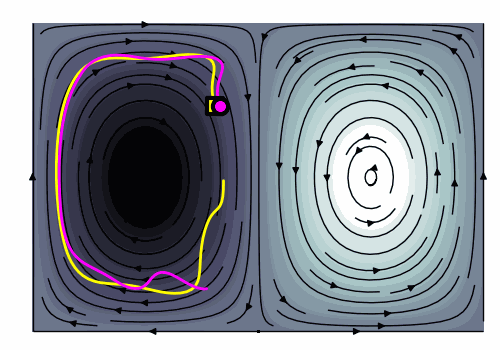}
      
    }
    \subfloat{
        \includegraphics[width=0.33\linewidth, height=0.2\linewidth]{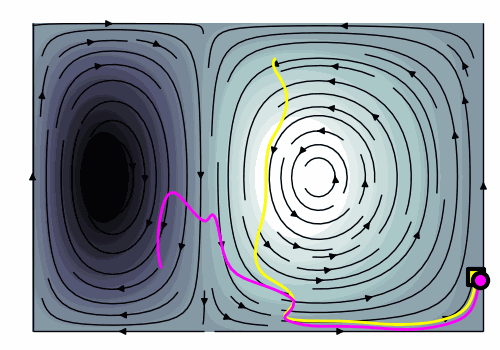}
      
    }
    \subfloat{
        \includegraphics[width=0.33\linewidth, height=0.2\linewidth]{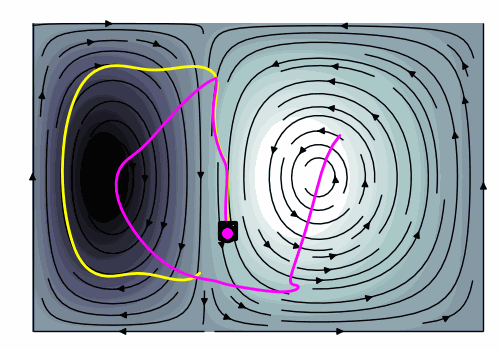}
      
    }

    \vspace{-0.5cm} 
    \begin{sideways}
    \makebox[0pt][l]{\hspace{-4.07cm}
    \begin{minipage}{4cm}{\large \bfseries 100 seconds} \end{minipage}}
    \end{sideways}\subfloat{
        \includegraphics[width=0.33\linewidth, height=0.2\linewidth]{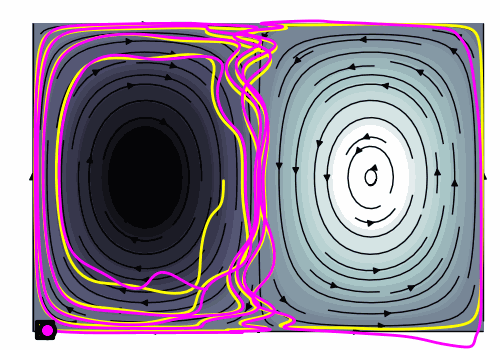}
      
    }
    \subfloat{
        \includegraphics[width=0.33\linewidth, height=0.2\linewidth]{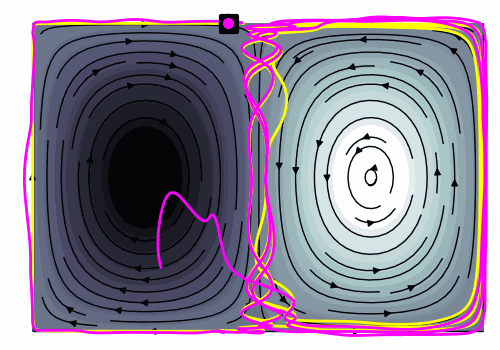}
      
    }
    \subfloat{
        \includegraphics[width=0.33\linewidth, height=0.2\linewidth]{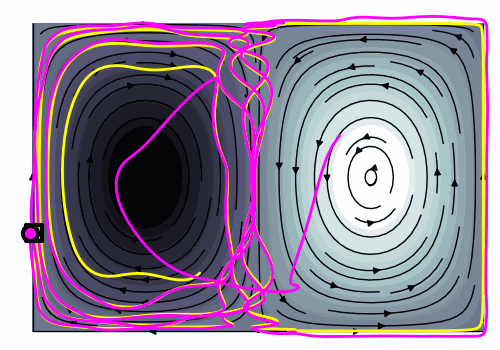}
      
    }

    \vspace{-0.25cm} 
    \subfloat{
        \includegraphics[width=0.335\linewidth]{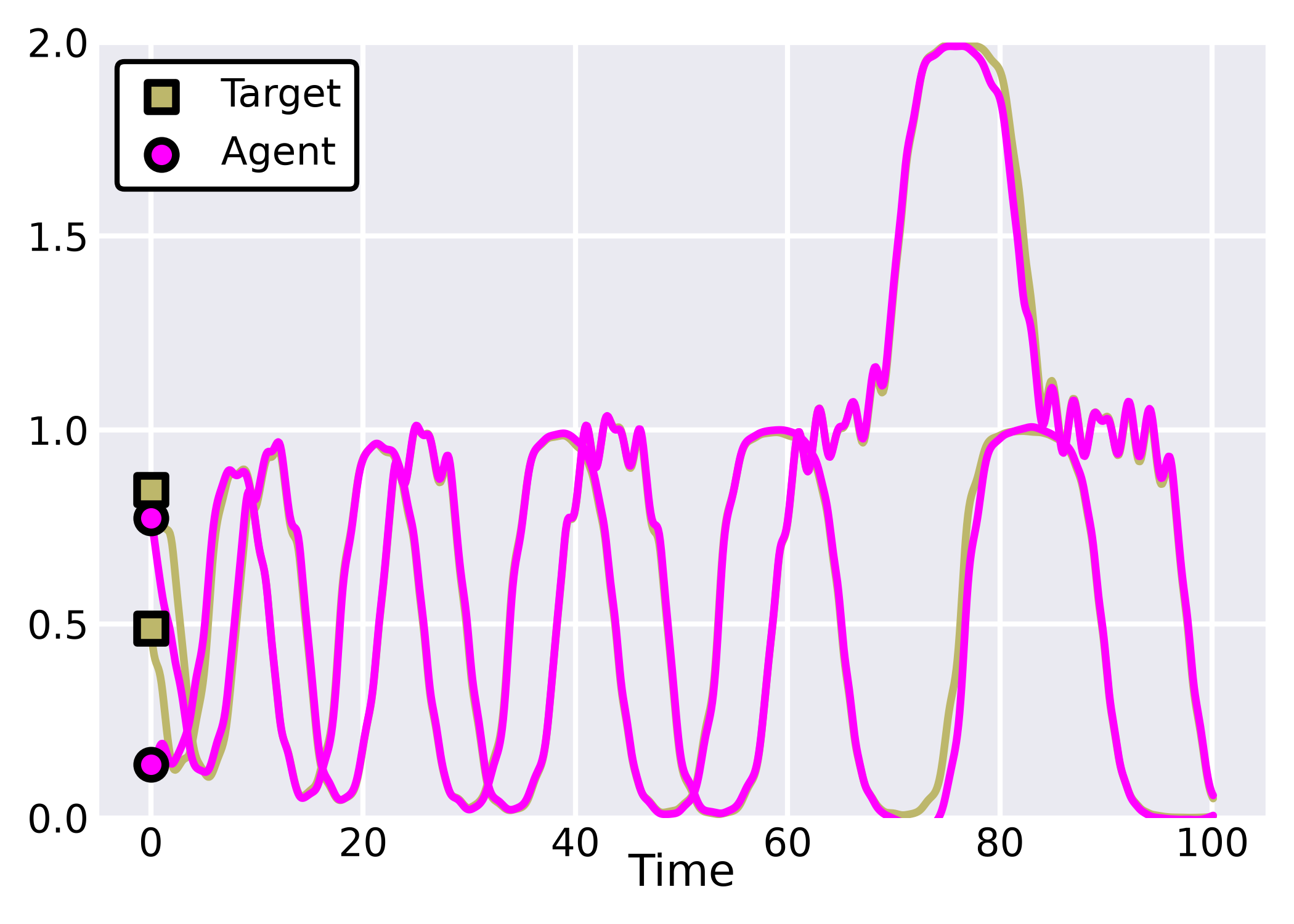}
      
    }
    \subfloat{
        \includegraphics[width=0.335\linewidth]{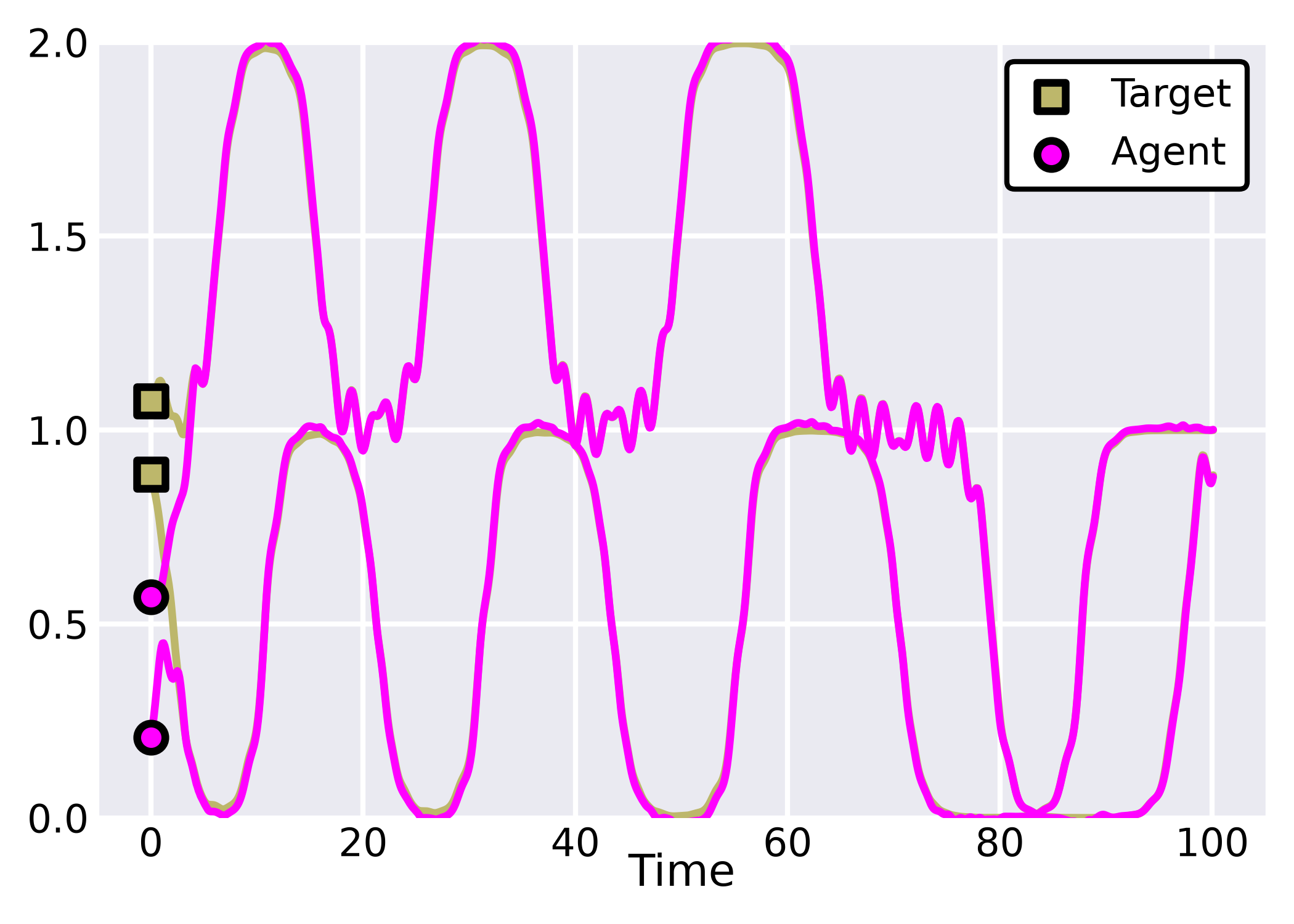}
      
    }
    \subfloat{
        \includegraphics[width=0.335\linewidth]{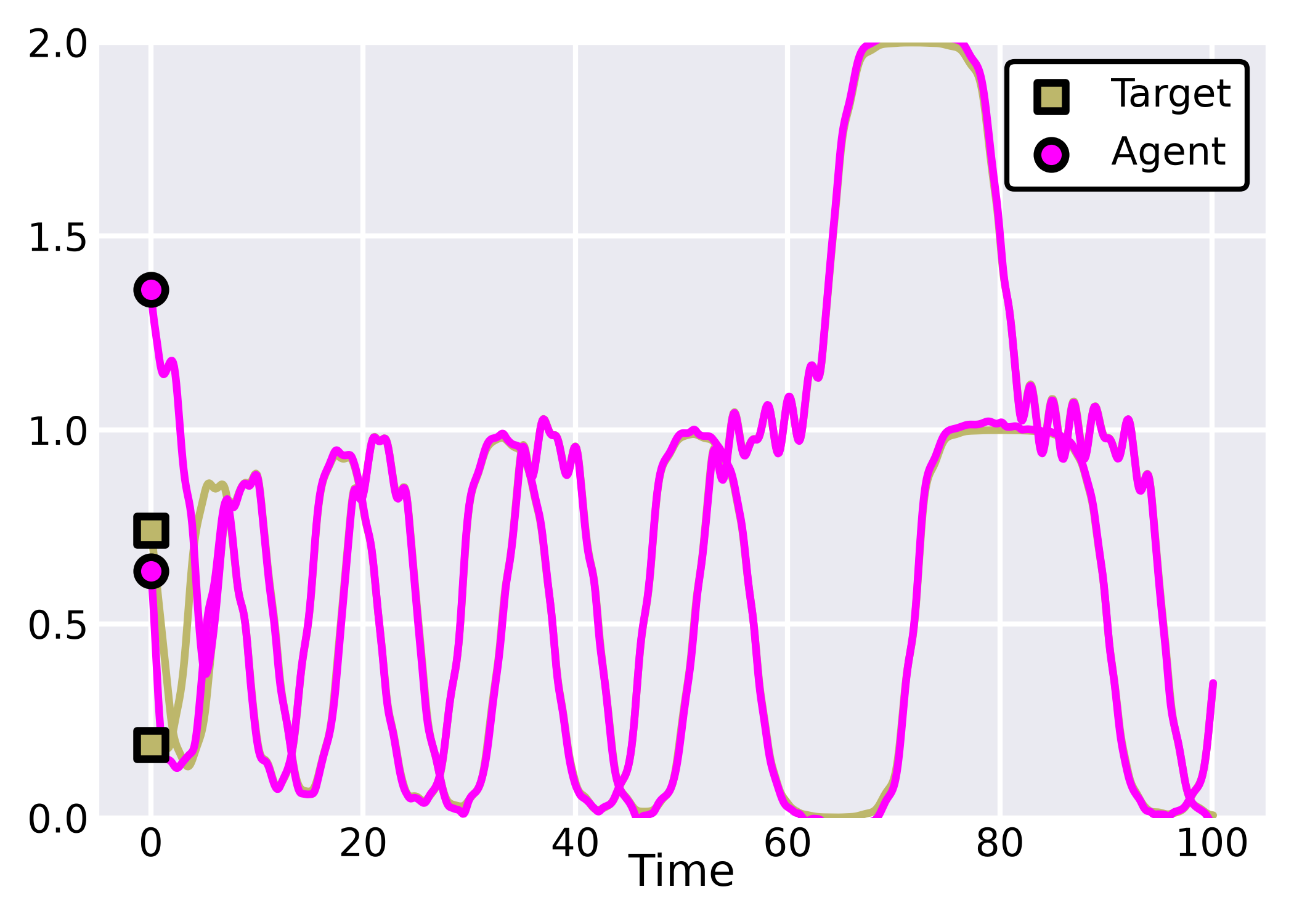}
      
    }

    \caption{{\em Leader-follower game}. Examples of different parametric controlled solutions in the leader-follower game using PEARL. The follower is indicated by the magenta circle and the leader by the yellow square. The first 4 rows show the state evolution over time, while the last shows the evolution of the leader and follower coordinates over time.}
    \label{fig:singleagent}
\end{figure*}

This first control problem involves an optimal navigation problem where an agent -- i.e., the follower -- has to track a target particle -- i.e., the leader. Both the leader and the follower are passively transported by an unsteady double gyre flow field, making the learning of the optimal policy extremely challenging due to the flow disturbances and the limited actuation capabilities. The dynamics of the agent position $\mathbf{x}_F(t)$ in the rectangular domain $\Omega = (0, 2) \times (0, 1)$ over the time interval $[0, 100]$ is defined as
\begin{equation}
\begin{cases}
\dfrac{d \mathbf{x}_F(t)}{dt} = \mathbf{v}(\mathbf{x}_F(t), t) + \mathbf{u}(t) \\
\mathbf{x}_F(0) = \mathbf{x}_{F,0} \\
\end{cases}
\label{eq:follower_double_gyre_controlled}
\end{equation}
where $\mathbf{v}$ denotes the double gyre flow velocity defined as 
\begin{equation}
\mathbf{v}(\mathbf{x}, t) = 
\begin{bmatrix}
-\pi A \sin\!\big(\pi f(x_1,t)\big)\cos(\pi x_2) \\
\pi A \cos\!\big(\pi f(x_1,t)\big)\sin(\pi x_2)\,\frac{\partial f(x_1,t)}{\partial x_1}
\end{bmatrix}
\label{eq:doublegyre}
\end{equation}
for every $\mathbf{x} = ({x}_1, {x}_2) \in \Omega$ and every $t \in [0,T]$, with $A=0.1$ being the intensity coefficient, $f(x_1,  t) = \epsilon \sin(\omega t) x_1^2 + (1 - 2\epsilon \sin(\omega t)) x_1$, while the perturbation amplitude and the frequency of oscillation are set equal to $\epsilon = 0.25$ and $\omega = \pi$, respectively. Moreover, $\mathbf{u}(t)$ represents the control velocity steering the agent, whose components are bounded in the interval $(-0.2, 0.2)$, while $\mathbf{x}_{F,0}$ represents the initial agent position uniformly sampled in the region $[0.1, 1.9] \times  [0.1, 0.9]$. 

The leader, instead, has no external actuation and it is solely transported by the gyre-field velocity, that is
\begin{equation}
\begin{cases}
\dfrac{d \mathbf{x}_L(t)}{dt} = \mathbf{v}(\mathbf{x}_L(t), t)  \\
\mathbf{x}_L(0) = \mathbf{x}_{L,0}, \\
\end{cases}
\label{eq:leader_double_gyre_controlled}
\end{equation}
where the initial position is uniformly sampled in the region $[0.1, 1.9] \times  [0.1, 0.9]$ at the beginning of each rollout. Importantly, the variability of the initial conditions of the leader and the follower requires parametric policies, capable of adapting to different scenarios. In this context, the follower position is available within the agent state, while the leader position is regarded as time-dependent scenario parameter $\boldsymbol{\mu}(t) = \mathbf{x}_{L}(t)$, thus resulting in time-varying inputs for the policy and the adjoint network. To solve the state equations for the leader and the follower in Equation~\eqref{eq:follower_double_gyre_controlled} and Equation~\eqref{eq:leader_double_gyre_controlled}, we consider a uniform grid of temporal instances $\{t_k\}_{k=0,\ldots,N_t}$ with $N_t = 1001$ and $\Delta t=0.1$ seconds, and we employ forward Euler, yielding the discrete-time equations
\begin{equation*}
\begin{split}
    \mathbf{x}_{F,k+1} &= \mathbf{x}_{F,k} + \Delta t (\mathbf{v}(\mathbf{x}_{F,k}, t_k) + \mathbf{u}_{k})
    \\
    \mathbf{x}_{L,k+1} &= \mathbf{x}_{L,k} + \Delta t \mathbf{v}(\mathbf{x}_{L,k}, t_k)
\end{split}
\end{equation*}
where $\mathbf{u}_k = \mathbf{u}(t_k)$, $\mathbf{x}_{F,k} = \mathbf{x}_{F}(t_k)$ and $\mathbf{x}_{L,k} = \mathbf{x}_{L}(t_k)$ for every $k=0,\ldots,N_t$. 

The goal of the agent is to reach and track the leader moving in the flow over time. To this aim, we investigate two possible reward functions. First, we consider the dense reward function $R_d$
\begin{equation}
    R_d(\mathbf{x}_{F,k}, \mathbf{u}_{k}, \boldsymbol{\mu}_k) = -\underbrace{||\mathbf{x}_{F,k} - \boldsymbol{\mu}_k||^2}_{\text{state term}} 
    -  \underbrace{\beta ||\mathbf{u}_{k}||^2}_{\text{action term}},
\label{eq:reward_function_leaderfollower}
\end{equation}
where $\boldsymbol{\mu}_k = \mathbf{x}_{L,k}$ for $k=0,\ldots,N_t$, and $\beta=0.2$ is a scalar coefficient balancing the contribution of state and action terms. Furthermore, we consider a sparse reward function $R_s$
\begin{equation*}
    R_s(\mathbf{x}_{F,k}, \mathbf{u}_{k}, \boldsymbol{\mu}_k) = \underbrace{100 e^{-100 ||\mathbf{x}_{F,k} - \boldsymbol{\mu}_k||^2 }}_{\text{state term}} 
    -  \underbrace{\beta ||\mathbf{u}_{k}||^2}_{\text{action term}},
\end{equation*}
assigning a bonus equal to $100$ at every time step where the follower is close to the leader. Note that, to preserve differentiability, we approximate the bonus via a scaled exponential function. The state of the agent is composed of the particle position $\mathbf{x}_{F,k}$ and the flow velocity $\mathbf{v}(\mathbf{x}_{F,k}, t_k)$ for every $k=0,\ldots,N_t$.

The neural networks into play -- that are the policy, the critic and the adjoint network, depending on the RL method -- are modeled with the double encoding described at the beginning of Section~\ref{sec:test}, with $64$ neurons. Moreover, we consider learning rates equal to $0.0001$, $h = 16$ steps in the short horizons, $1500$ episodes of training and $10$ episodes of evaluation. Figure~\ref{fig:singleagent_rew} shows the training and evaluation rewards achieved by different agents over training and evaluation when employing the dense and the sparse reward functions. In both cases, PEARL is able to achieve the best rewards, outperforming not only the model-free baselines PPO and TD3, but also the gradient-based ones, namely BPTT, truncated BPTT, and SHAC. Due to the length of the control horizon, BPTT does not achieve good results due to exploding gradients which compromise optimization. Truncated BPTT improves the BPTT results, but it is not capable of retrieving robust optimal policies due to the myopic reward maximization over short horizons. On the other side, the model-free baselines, while achieving sufficient performance with dense rewards, are not able to learn acceptable policies when the rewards are sparse. Despite the similar working principles, PEARL outperforms SHAC, showing that learning the value gradients directly is more effective that learning the values, especially in the case of sparse rewards.

Eventually, Figure \ref{fig:singleagent} shows three examples of controlled solution in the leader-follow game when using PEARL. PEARL is capable of successfully and effectively controlling the follower to track the moving target in a double gyre flow, making their trajectories almost overlapping after few seconds of simulation. 

\subsection{MEAN-FIELD LEADER-FOLLOWER GAME}
\label{subsec:mean-field_lfg}

In this section we consider a high-dimensional variation of the leader-follower game in a double-gyre flow. The leader dynamics is the same as the one introduced in Section~\ref{subsec:lfg}, with initial position regarded as time-dependent scenario parameters $\boldsymbol{\mu}$. The agent, instead, is a density defined over the whole rectangular domain $\Omega$. Note that the mean-field description may be helpful to model pollutant concentration, as well as swarms at the macroscopic level. Specifically, starting from a Gaussian density $y_0$ centered at $(y_{0,1},y_{0,2})$ with variance $0.05$, the density dynamics is defined with the diffusion-advection PDE
\begin{equation}
\begin{cases}
\dfrac{dy}{dt} + \nabla \cdot (- \nu 
\nabla y + \mathbf{v} y + \mathbf{u} y) = 0 \qquad &\mbox{in} \, \Omega \times (0,T]
\\
(- \nu \nabla y + \mathbf{v}y + \mathbf{u}y) \cdot \mathbf{n} = 0 \qquad &\mbox{on} \, \partial \Omega  \times (0,T]
\\
y = y_0 \qquad &\mbox{in} \, \Omega \times \{t=0\} ,
\end{cases}
\label{eq:fp}
\end{equation}
where $\nu = 0.001$ is the diffusivity, $\mathbf{n}$ is the normal unit vector, and homogeneous Neumann boundary conditions are taken into account on the domain boundary $\partial \Omega$. Similarly to the previous test case, $\mathbf{v}(\mathbf{x},t)$ represent the double gyre flow velocity, while $\mathbf{u}(\mathbf{x},t)$ is the distributed control action defined over the whole space-time domain.  Regarding the time discretization, we consider a uniform grid within the horizon $[0,T]$, with final time $T=20$ seconds and time step $\Delta t=0.2$ seconds, that is $N_t=100$. Moreover, we spatially discretize the problem with finite elements, ending up with $N_y = 2145$ degrees of freedom for the discrete states $\mathbf{y}_k$, and $N_u = 4290$ for the discrete control action $\mathbf{u}_k$ for $k=0,\ldots,N_t$.

The goal is to steer the density towards the moving target for different initial mean position of the density $(y_{0,1},y_{0,2})$ in the region $[0.3, 1.7] \times [0.3, 0.7]$ and different initial position of the leader in the range $[0.1, 1.9] \times [0.1, 0.9]$. To do so, we aim at maximizing the reward function at the continuous level
\begin{equation*}
\begin{aligned}
R(y_k, u_k, \boldsymbol{\mu}_k) = & - 0.5 \int_\Omega (y_k - \bar{y}_k)^2 d\mathbf{x} - 0.5 \int_{\partial \Omega} y_k^2 d\Gamma - \\
& -0.5 \beta \int_\Omega ||u_k||^2 d\mathbf{x} - \beta_g \int_{\Omega} ||\nabla u_k ||^2 dx,
\end{aligned}
\label{eq:rew_meanfield}
\end{equation*}
where $\beta = \beta_g = 0.1$ balance state and control terms. 

\begin{figure*}[t]
    \centering \subfloat{\includegraphics[width=0.5\linewidth]{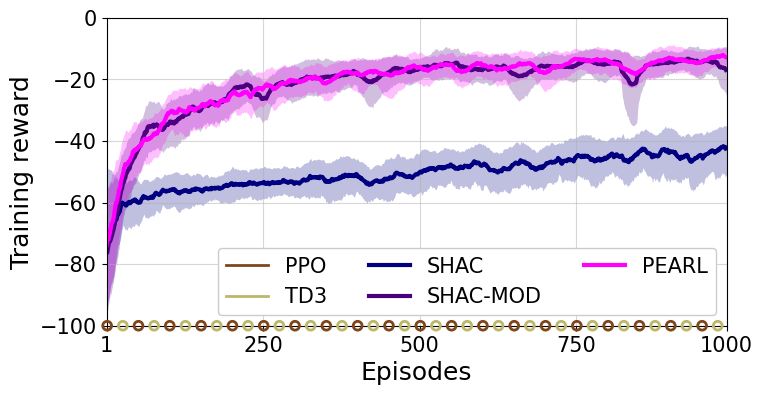}} \subfloat{\includegraphics[width=0.5\linewidth]{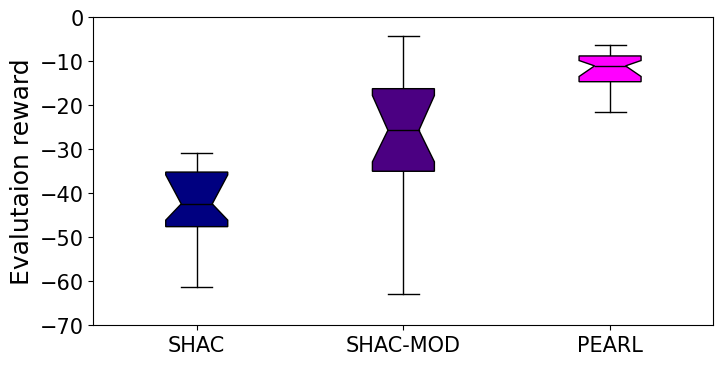}}
    
    \caption{{\em Mean-field leader-follower game}. Training and evaluation rewards obtained by the different competing agents in the leader-follower game with dense and sparse rewards.}
    \label{fig:meanfield_rew}
\end{figure*}

The neural networks of the competing methods, i.e., PPO, TD3, PEARL and SHAC-MOD, are modeled with the double encoding described at the beginning of Section~\ref{sec:test}, with $1024$ neurons per layer. To emphasize the importance of the proposed neural network architecture with double encoding, we test the original implementation of SHAC considering a single feed forward neural network with $4$ layers of $1024$ neurons. Moreover, we consider learning rates equal to 0.00001,
$h = 16$ steps in the short horizons, $1000$ episodes of training and $10$ episodes of evaluation. Figure~\ref{fig:meanfield_rew} shows the training and evaluation rewards achieved by different agents over training and evaluation when employing the dense reward function in Equation \eqref{eq:rew_meanfield}. PEARL and SHAC-MOD are able to achieve the best rewards over training, yet PEARL is the best-performing agent during evaluation. SHAC relying on the original neural network architecture proposed in \cite{xu2022accelerated} is not able to achieve rewards comparable to PEARL. In this high-dimensional control problem, instead, the model-free RL baselines (PPO and TD3) are not capable of learning any meaningful policy, experiencing exploding gradients due to the stochastic exploration policy compromising the numerical stability of the solver.

\begin{figure*}[t]
    \centering
    \begin{sideways}
    \makebox[0pt][l]{\hspace{-3.5cm}
    \begin{minipage}{4cm}{\large \bfseries 0 seconds} \end{minipage}}
    \end{sideways}\hspace{0.2cm}\subfloat[\large \bfseries Test 1]{
        \includegraphics[width=0.3\linewidth]{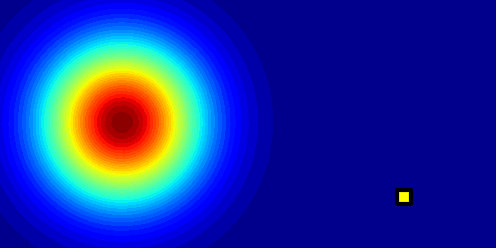}
      
    }
    \subfloat[\large \bfseries Test 2]{
        \includegraphics[width=0.3\linewidth]{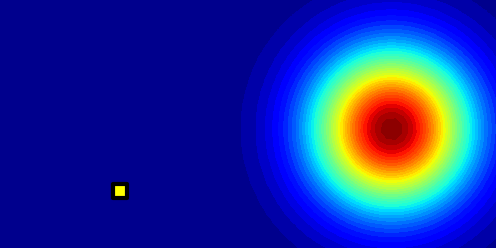}
      
    }
    \subfloat[\large \bfseries Test 3]{
        \includegraphics[width=0.3\linewidth]{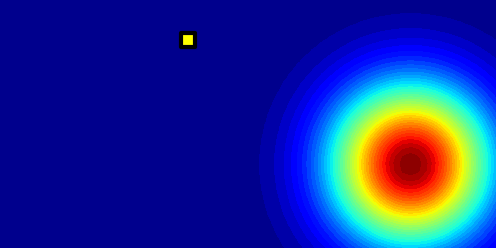}
      
    }

    \vspace{-0.25cm} 
    \begin{sideways}
    \makebox[0pt][l]{\hspace{-3.5cm}
    \begin{minipage}{4cm}{\large \bfseries 1 second} \end{minipage}}
    \end{sideways}\hspace{0.2cm}\subfloat{
        \includegraphics[width=0.3\linewidth]{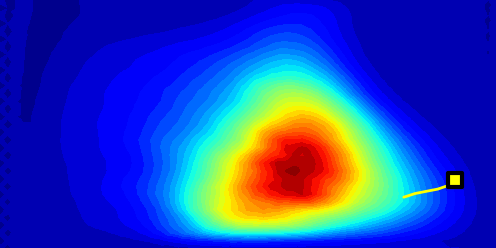}
      
    }
    \subfloat{
        \includegraphics[width=0.3\linewidth]{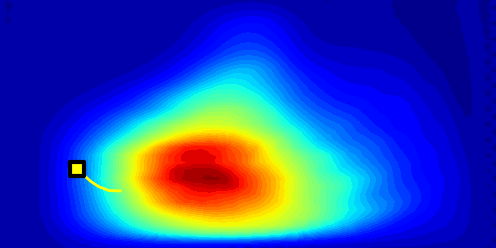}
      
    }
    \subfloat{
        \includegraphics[width=0.3\linewidth]{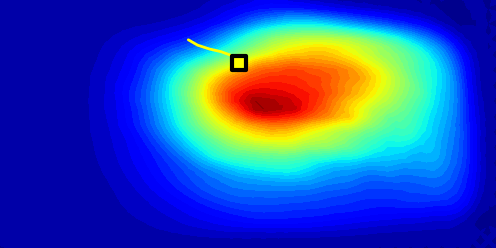}
      
    }

    \vspace{-0.25cm} 
    \begin{sideways}
    \makebox[0pt][l]{\hspace{-3.5cm}
    \begin{minipage}{4cm}{\large \bfseries 2 seconds} \end{minipage}}
    \end{sideways}\hspace{0.2cm}\subfloat{
        \includegraphics[width=0.3\linewidth]{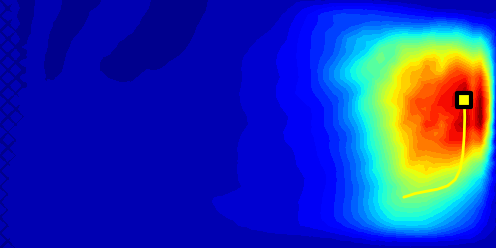}
      
    }
    \subfloat{
        \includegraphics[width=0.3\linewidth]{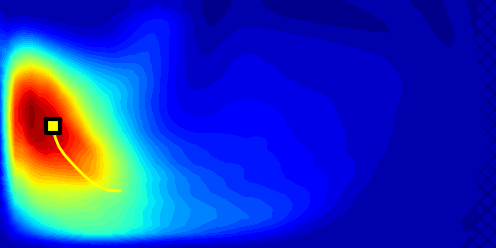}
      
    }
    \subfloat{
        \includegraphics[width=0.3\linewidth]{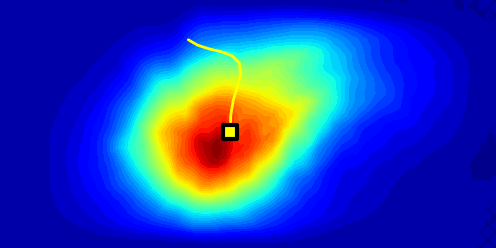}
      
    }

    \vspace{-0.25cm} 
    \begin{sideways}
    \makebox[0pt][l]{\hspace{-3.5cm}
    \begin{minipage}{4cm}{\large \bfseries 5 seconds} \end{minipage}}
    \end{sideways}\hspace{0.2cm}\subfloat{
        \includegraphics[width=0.3\linewidth]{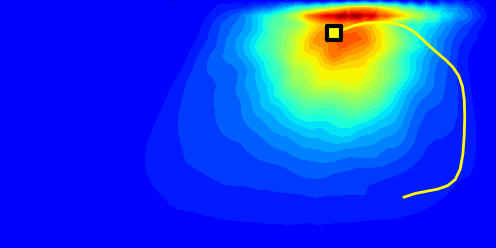}
      
    }
    \subfloat{
        \includegraphics[width=0.3\linewidth]{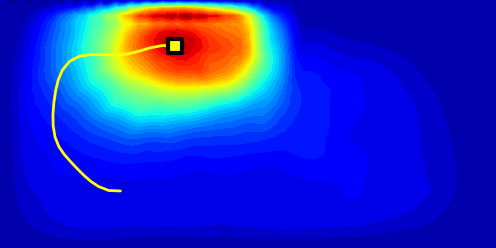}
      
    }
    \subfloat{
        \includegraphics[width=0.3\linewidth]{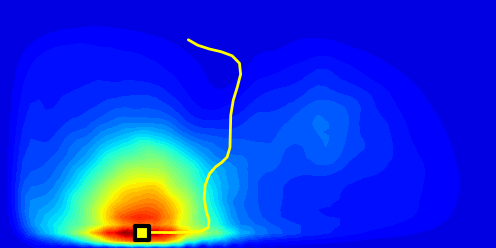}
      
    }

    \vspace{-0.25cm} 
    \begin{sideways}
    \makebox[0pt][l]{\hspace{-3.5cm}
    \begin{minipage}{4cm}{\large \bfseries 10 seconds} \end{minipage}}
    \end{sideways}\hspace{0.2cm}\subfloat{
        \includegraphics[width=0.3\linewidth]{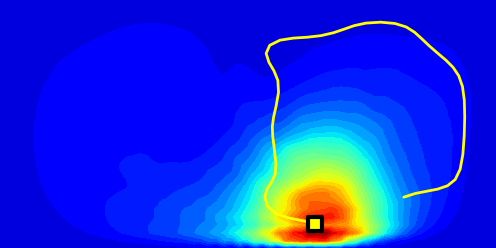}
      
    }
    \subfloat{
        \includegraphics[width=0.3\linewidth]{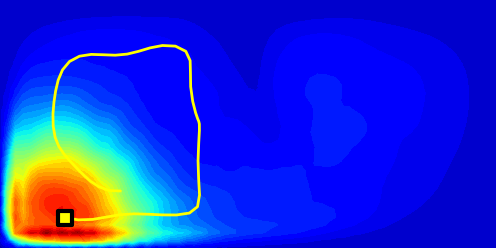}
      
    }
    \subfloat{
        \includegraphics[width=0.3\linewidth]{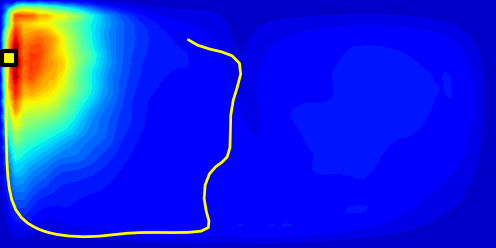}
      
    }

    \vspace{-0.25cm} 
    \begin{sideways}
    \makebox[0pt][l]{\hspace{-3.5cm}
    \begin{minipage}{4cm}{\large \bfseries 20 seconds} \end{minipage}}
    \end{sideways}\hspace{0.2cm}\subfloat{
        \includegraphics[width=0.3\linewidth]{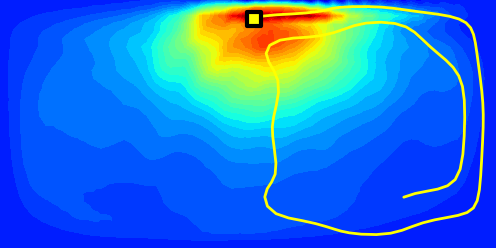}
      
    }
    \subfloat{
        \includegraphics[width=0.3\linewidth]{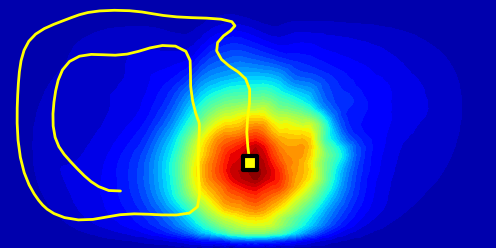}
      
    }
    \subfloat{
        \includegraphics[width=0.3\linewidth]{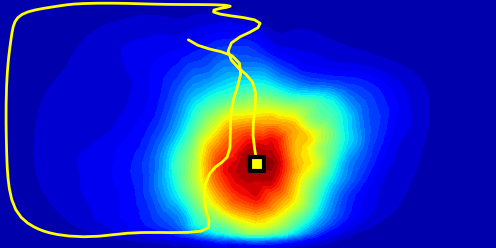}
      
    }
    \vspace{0.25cm}
    \subfloat{\includegraphics[scale=0.38]{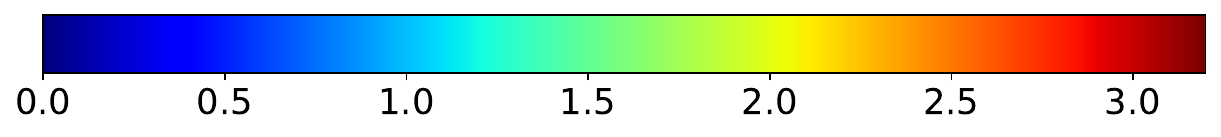}}

    \caption{{\em Mean-field leader-follower game}. Examples of different parametric controlled solutions in the mean-field leader-follower game using PEARL. The leader is denoted by the yellow square. The different rows show the state evolution over time.}
        \label{fig:meanfield_states}
\end{figure*}

\begin{figure*}[t]
    \centering
    \begin{sideways}
    \makebox[0pt][l]{\hspace{-3.5cm}
    \begin{minipage}{4cm}{\large \bfseries 0 seconds} \end{minipage}}
    \end{sideways}\hspace{0.2cm}\subfloat[\large \bfseries Test 1]{
        \includegraphics[width=0.3\linewidth]{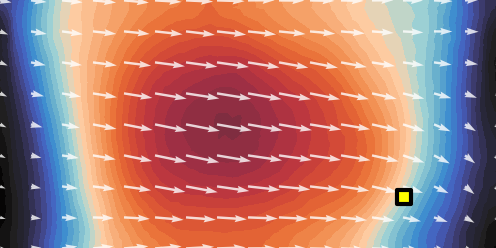}
      
    }
    \subfloat[\large \bfseries Test 2]{
        \includegraphics[width=0.3\linewidth]{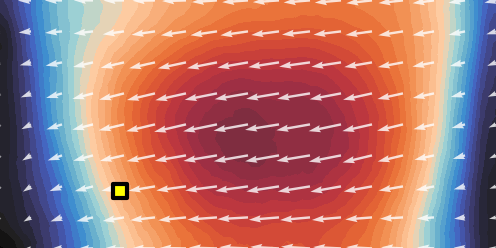}
      
    }
    \subfloat[\large \bfseries Test 3]{
        \includegraphics[width=0.3\linewidth]{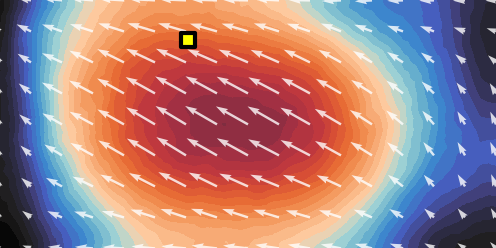}
      
    }

    \vspace{-0.25cm} 
    \begin{sideways}
    \makebox[0pt][l]{\hspace{-3.5cm}
    \begin{minipage}{4cm}{\large \bfseries 1 second} \end{minipage}}
    \end{sideways}\hspace{0.2cm}\subfloat{
        \includegraphics[width=0.3\linewidth]{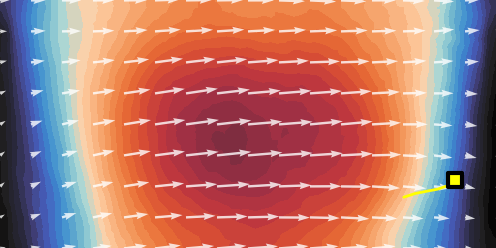}
      
    }
    \subfloat{
        \includegraphics[width=0.3\linewidth]{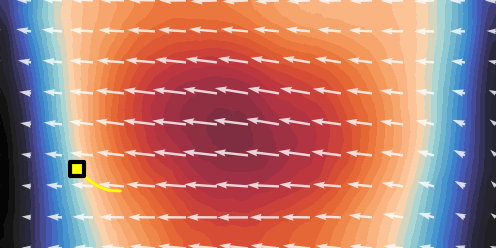}
      
    }
    \subfloat{
        \includegraphics[width=0.3\linewidth]{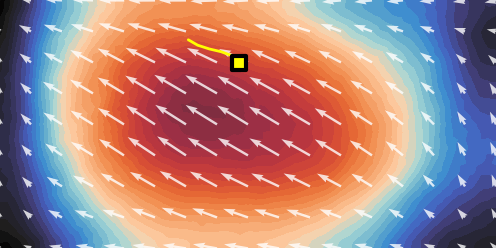}
      
    }

    \vspace{-0.25cm} 
    \begin{sideways}
    \makebox[0pt][l]{\hspace{-3.5cm}
    \begin{minipage}{4cm}{\large \bfseries 2 seconds} \end{minipage}}
    \end{sideways}\hspace{0.2cm}\subfloat{
        \includegraphics[width=0.3\linewidth]{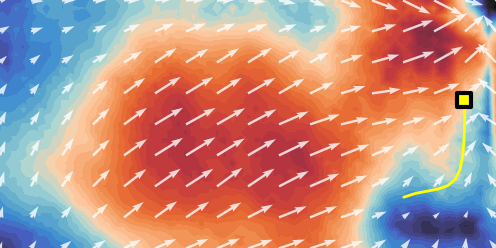}
      
    }
    \subfloat{
        \includegraphics[width=0.3\linewidth]{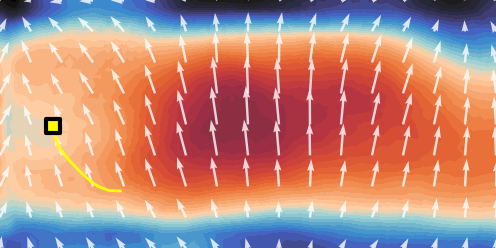}
      
    }
    \subfloat{
        \includegraphics[width=0.3\linewidth]{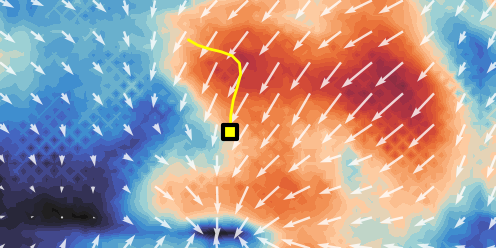}
      
    }

    \vspace{-0.25cm} 
    \begin{sideways}
    \makebox[0pt][l]{\hspace{-3.5cm}
    \begin{minipage}{4cm}{\large \bfseries 5 seconds} \end{minipage}}
    \end{sideways}\hspace{0.2cm}\subfloat{
        \includegraphics[width=0.3\linewidth]{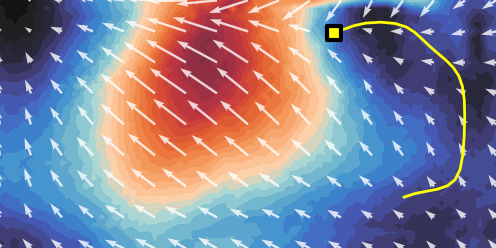}
      
    }
    \subfloat{
        \includegraphics[width=0.3\linewidth]{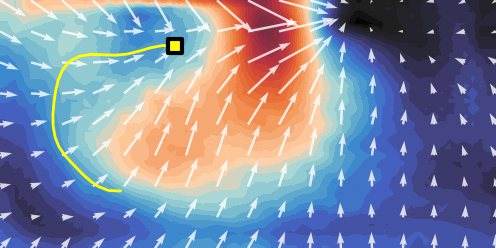}
      
    }
    \subfloat{
        \includegraphics[width=0.3\linewidth]{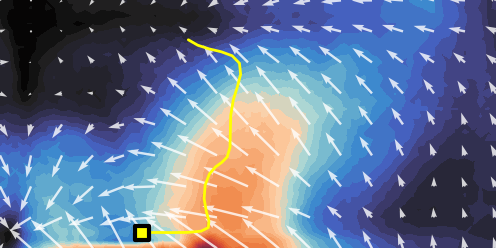}
      
    }

    \vspace{-0.25cm} 
    \begin{sideways}
    \makebox[0pt][l]{\hspace{-3.5cm}
    \begin{minipage}{4cm}{\large \bfseries 10 seconds} \end{minipage}}
    \end{sideways}\hspace{0.2cm}\subfloat{
        \includegraphics[width=0.3\linewidth]{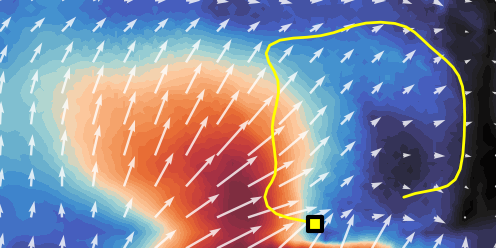}
      
    }
    \subfloat{
        \includegraphics[width=0.3\linewidth]{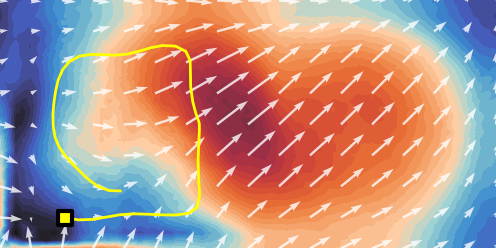}
      
    }
    \subfloat{
        \includegraphics[width=0.3\linewidth]{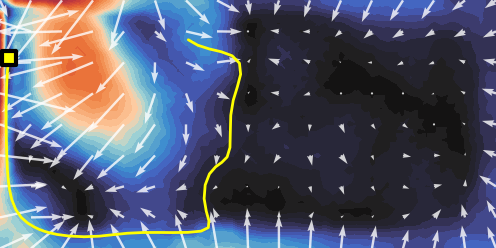}
      
    }

    \vspace{-0.25cm} 
    \begin{sideways}
    \makebox[0pt][l]{\hspace{-3.5cm}
    \begin{minipage}{4cm}{\large \bfseries 20 seconds} \end{minipage}}
    \end{sideways}\hspace{0.2cm}\subfloat{
        \includegraphics[width=0.3\linewidth]{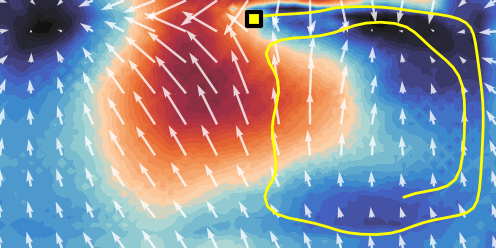}
      
    }
    \subfloat{
        \includegraphics[width=0.3\linewidth]{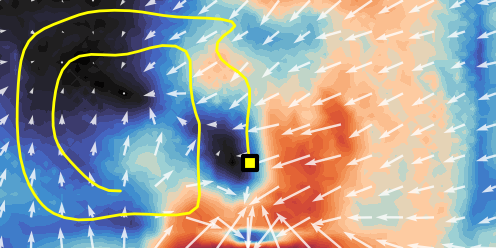}
      
    }
    \subfloat{
        \includegraphics[width=0.3\linewidth]{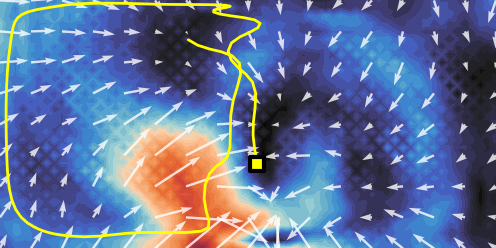}
      
    }
    \vspace{0.25cm}
    \subfloat{\includegraphics[scale=0.38]{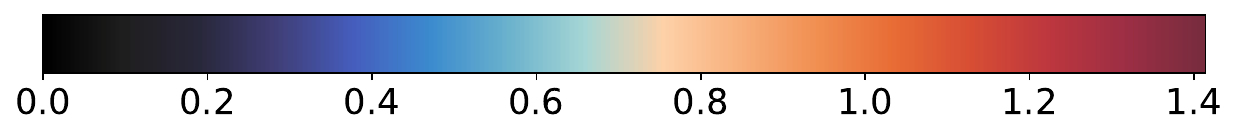}}

    \caption{{\em Mean-field leader-follower game}. Examples of different control actions in the mean-field leader-follower game using PEARL. The leader is denoted by the yellow square. The different rows show the action evolution over time.}
        \label{fig:meanfield_controls}
\end{figure*}

\begin{figure*}[t]
    \centering
    \begin{sideways}
    \makebox[0pt][l]{\hspace{-3.5cm}
    \begin{minipage}{4cm}{\large \bfseries Episode 1} \end{minipage}}
    \end{sideways}\hspace{0.2cm}\subfloat[\shortstack{\large \bfseries PPO \\ \large Control -- 10 seconds}]{
        \includegraphics[width=0.3\linewidth]{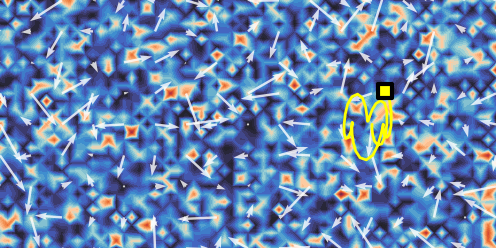}
      
    }
    \subfloat[\shortstack{\large \bfseries PEARL \\ \large Control -- 10 seconds}]{
        \includegraphics[width=0.3\linewidth]{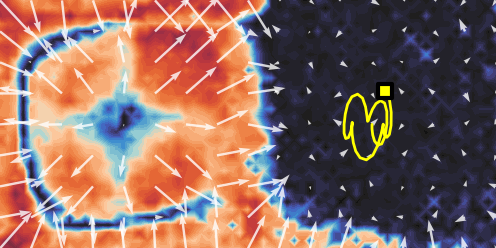}
      
    }
    \subfloat[\shortstack{\large \bfseries PEARL \\ \large State -- 10 seconds}]{
        \includegraphics[width=0.3\linewidth]{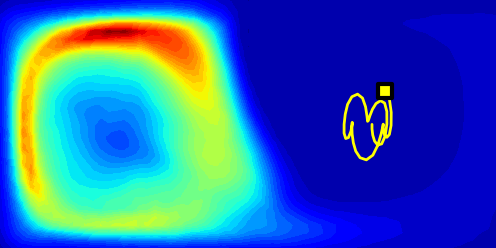}
      
    }

    \vspace{-0.25cm} 
    \begin{sideways}
    \makebox[0pt][l]{\hspace{-3.5cm}
    \begin{minipage}{4cm}{\large \bfseries Episode 2} \end{minipage}}
    \end{sideways}\hspace{0.2cm}\subfloat{
        \includegraphics[width=0.3\linewidth]{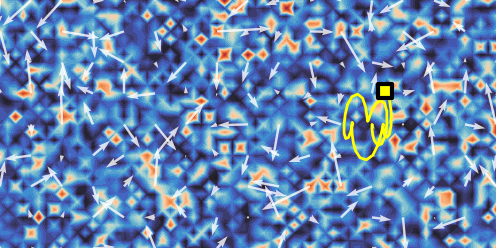}
      
    }
    \subfloat{
        \includegraphics[width=0.3\linewidth]{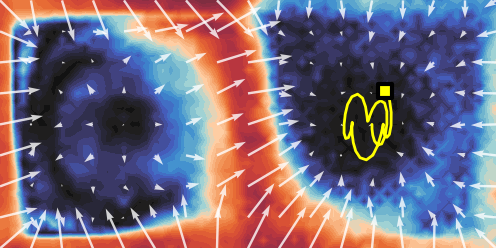}
      
    }
    \subfloat{
        \includegraphics[width=0.3\linewidth]{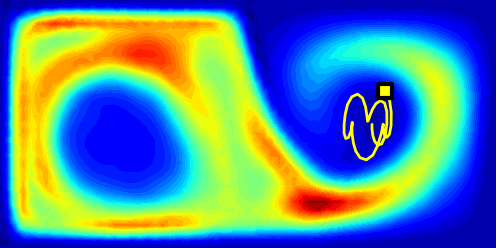}
      
    }

    \vspace{-0.25cm} 
    \begin{sideways}
    \makebox[0pt][l]{\hspace{-3.5cm}
    \begin{minipage}{4cm}{\large \bfseries Episode 3} \end{minipage}}
    \end{sideways}\hspace{0.2cm}\subfloat{
        \includegraphics[width=0.3\linewidth]{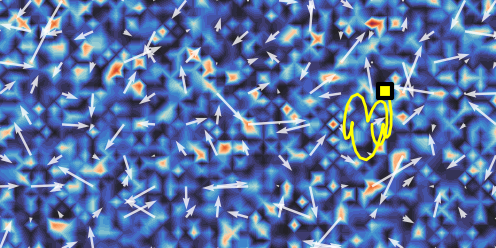}
      
    }
    \subfloat{
        \includegraphics[width=0.3\linewidth]{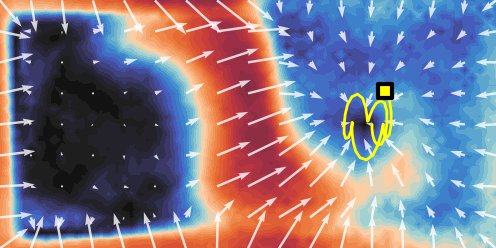}
      
    }
    \subfloat{
        \includegraphics[width=0.3\linewidth]{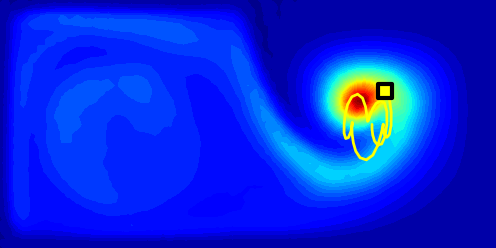}
      
    }

    \vspace{-0.25cm} 
    \begin{sideways}
    \makebox[0pt][l]{\hspace{-3.5cm}
    \begin{minipage}{4cm}{\large \bfseries Episode 5} \end{minipage}}
    \end{sideways}\hspace{0.2cm}\subfloat{
        \includegraphics[width=0.3\linewidth]{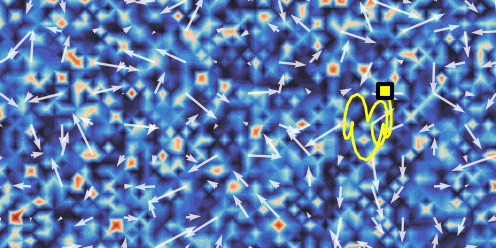}
      
    }
    \subfloat{
        \includegraphics[width=0.3\linewidth]{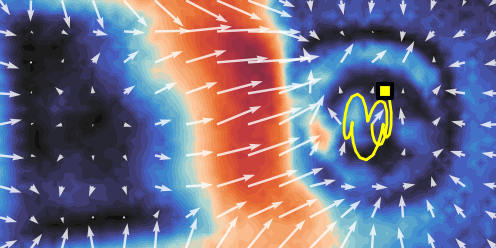}
      
    }
    \subfloat{
        \includegraphics[width=0.3\linewidth]{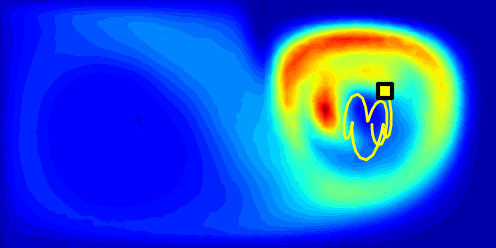}
      
    }

    \vspace{-0.25cm} 
    \begin{sideways}
    \makebox[0pt][l]{\hspace{-3.5cm}
    \begin{minipage}{4cm}{\large \bfseries Episode 10} \end{minipage}}
    \end{sideways}\hspace{0.2cm}\subfloat{
        \includegraphics[width=0.3\linewidth]{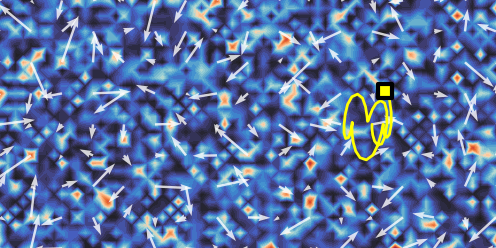}
      
    }
    \subfloat{
        \includegraphics[width=0.3\linewidth]{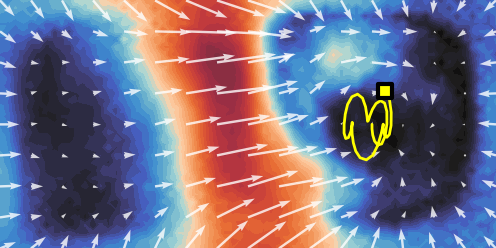}
      
    }
    \subfloat{
        \includegraphics[width=0.3\linewidth]{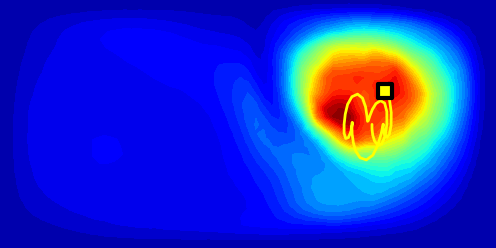}
      
    }

    \vspace{-0.25cm} 
    \begin{sideways}
    \makebox[0pt][l]{\hspace{-3.5cm}
    \begin{minipage}{4cm}{\large \bfseries Evaluation} \end{minipage}}
    \end{sideways}\hspace{0.3cm}\subfloat{
        \includegraphics[width=0.3\linewidth]{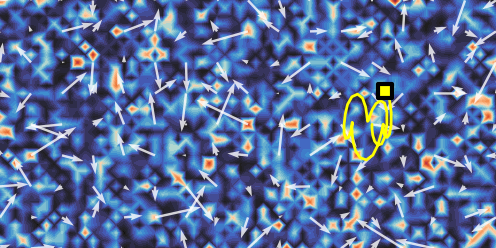}
      
    }
    \subfloat{
        \includegraphics[width=0.3\linewidth]{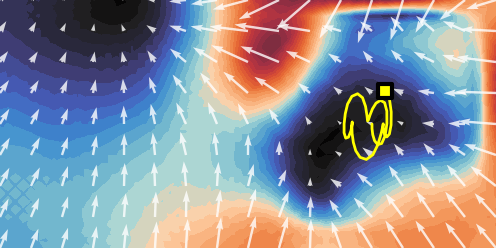}
      
    }
    \subfloat{
        \includegraphics[width=0.3\linewidth]{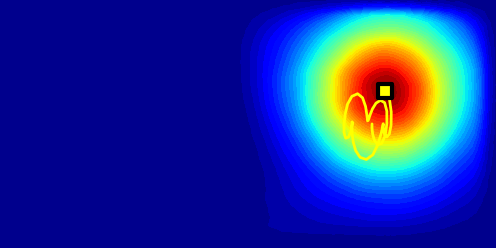}
      
    }
    
    \vspace{0.25cm}
    \begin{sideways}
    \makebox[0pt][l]{\hspace{-3.5cm}
    \begin{minipage}{4cm}{\hphantom{\large \bfseries Evaluation}} \end{minipage}}
    \end{sideways}\hspace{0.1cm} \subfloat{\includegraphics[scale=0.38]{images/meanfield_leaderfollower/control_cmap.pdf}} \quad \subfloat{\includegraphics[scale=0.38]{images/meanfield_leaderfollower/state_cmap.pdf}}

    \caption{{\em Mean-field leader-follower game}. Comparison between the controls of PEARL and PPO at several stages of the training. It is worth noticing that after the first updates the control actions learn by PEARL are smooth and capable of preventing the explosion of the numerical solver, and, consequently, of the rewards.}
        \label{fig:meanfield_training}
\end{figure*}

In Figure \ref{fig:meanfield_states} and \ref{fig:meanfield_controls} we show three examples of controlled solution in the mean-field leader-follow game when using PEARL. Despite the high-dimensionality of the problem, PEARL is capable of controlling the density to track the target moving in a double gyre flow. Eventually, in Figure \ref{fig:meanfield_training} we show a comparison between the controls of PEARL and PPO at different stages of the training. It is worth noticing that after the first updates the control actions learn by PEARL are smooth and capable of preventing the explosion of the numerical solver, and, consequently, of the rewards, leading to the efficient learning of the optimal policy.

\section{CONCLUSIONS}
\label{sec:conlusions}

In this work, we bridge reinforcement learning and traditional optimal control theory for the closed-loop control of differentiable dynamical systems. After highlighting connections between the two frameworks in Section~\ref{sec:rlocp}, as well as pros and cons, we propose a novel {\em Physics-EnhAnced Reinforcement Learning} (PEARL) algorithm tailored to control of high-dimensional, distributed and parametrized differentiable environments. Unlike classical model-free RL approaches, PEARL directly exploits the governing physics of the environment to compute policy gradients via automatic differentiation, speeding up the optimization, and tackling the curse of dimensionality. 
By combining short-horizon optimization, automatic differentiation, and a neural network-based approximation of the terminal adjoint of each short horizon to account for long-term dependencies, PEARL mitigates the vanishing or exploding gradients that plague backpropagation through time over long rollouts and prevent the learning of myopic contorl policies that affects truncated backpropagation through time, without sacrificing the closed-loop, real-time applicability that classical optimal control methods typically lack. 
Importantly, we exploit the adjoint equation to generate target data for adjoint network training, thus providing more informative and effective gradients with respect to value learning methods. 

Through two challenging parametric navigation problems in unsteady double-gyre flows -- a leader-follower game and its mean-field, high-dimensional counterpart -- we demonstrated that PEARL outperforms state-of-the-art model-free algorithms, such as PPO and TD3, as well as AD-based alternatives, such as BPTT, truncated BPTT and SHAC. Notably, PEARL achieves outstanding performance without resorting to low-dimensional state representations or multi-agent decompositions, directly handling high-dimensional states and distributed actions, and generalizes across multiple scenario parameters within a single policy. These results indicate that embedding adjoint-based sensitivities directly into the policy gradient offers a principled and effective route to sample-efficient, physics-consistent policy learning. 

As future work, we will apply PEARL to larger-scale and more complex control problems, for example, arising in active flow control. In addition, we will extend our algorithm to deal with partially observable dynamics, in the direction of real-world settings. Moreover, model-based extensions may be developed with differentiable surrogate models emulating the real environment dynamics in order to further enhance sample efficiency.
         
\section*{CODE AVAILABILITY} 
The code can be found in our repository \\
\url{https://github.com/MatteoTomasetto/pearl}.

\section*{ACKNOWLEDGEMENTS} 
NB and AM acknowledge the Project “Reduced Order Modeling and Deep Learning for the real-time approximation of PDEs (DREAM)” (Starting Grant No. FIS00003154), funded by the Italian Science Fund (FIS) - Ministero dell'Università e della Ricerca. AM also acknowledges the project “Dipartimento di Eccellenza” 2023-2027 funded by MUR.

\vfill
\onecolumngrid
\vspace{5mm}
\noindent\centering\rule{0.5\textwidth}{1pt}
\vspace{5mm}

\twocolumngrid
\renewcommand{\bibsection}{}
\bibliographystyle{abbrv}
\bibliography{references}

\vfill
\onecolumngrid
\vspace{5mm}
\noindent\centering\rule{0.5\textwidth}{1pt}
\vspace{5mm}

\onecolumngrid
\raggedright
\makeatletter
\let\centering\raggedright
\makeatother
\appendix

\section{Sample complexity of inexact gradient algorithms}
\label{sec:appendix1}

In this section, we derive the sample complexity of the gradient ascent algorithm when the ascent direction is approximated through a stochastic estimator. 
Let us consider the gradient ascent algorithm in the policy parameter space
\begin{equation}
    \boldsymbol{\theta}^{(n+1)} = \boldsymbol{\theta}^{(n)} + \eta \hat{g}_n \; \text{ for } \, n = 1,\ldots,N
    \label{eq:update}
\end{equation}
where $\eta >0$ is the learning rate, $N > 0$ is the number of iterations, while $\hat{g}_n \in \mathbb{R}^{N_\theta}$ is an unbiased estimator of the policy gradient $\nabla_{\boldsymbol{\theta}} G(\boldsymbol{\theta}^{(n)})$ for every iteration $n=1,\ldots,N$, i.e.  $\hat{g}_n = \nabla_{\boldsymbol{\theta}} G(\boldsymbol{\theta}^{(n)}) + \boldsymbol{\xi}_n$ with $\mathbb{E}[\boldsymbol{\xi}_n]=\boldsymbol{0} \in \mathbb{R}^{N_\theta}$. Moreover, let us assume that the gradient estimate has bounded variance, that is $\mathbb{E}[\lVert \boldsymbol{\xi}_n\rVert^2]\leq \sigma^2$ for every iteration $n=1,\ldots,N$, where $\lVert \cdot \rVert$ denotes the Euclidean norm. Notice that, while throughout this work we consider gradients as row vectors, in this section they are regarded as column vectors to simplify the notation. Consistently with \cite{nesterov2013introductory, Nesterov2015}, we assume that $G$ is L-smooth, that is $\lVert \nabla_{\boldsymbol{\theta}} G(\boldsymbol{\theta}_1) - \nabla_{\boldsymbol{\theta}} G(\boldsymbol{\theta}_2)\rVert \leq L \lVert\boldsymbol{\theta}_1-\boldsymbol{\theta}_2\rVert$ for all $\boldsymbol{\theta}_1,\boldsymbol{\theta}_2 \in \mathbb{R}^{N_\theta}$ with constant $L>0$. Notice that the negative of $G$ is also L-smooth. The L-smoothness of the negative of $G$ is equivalent to~\cite{nesterov2013introductory, Nesterov2015} 
\begin{equation*}
G(\boldsymbol{\theta}^{(n+1)}) \geq G(\boldsymbol{\theta}^{(n)}) +  \nabla_{\boldsymbol{\theta}} G(\boldsymbol{\theta}^{(n)})^\top( \boldsymbol{\theta}^{(n+1)}-\boldsymbol{\theta}^{(n)}) - \dfrac{L}{2}\lVert \boldsymbol{\theta}^{(n+1)}-\boldsymbol{\theta}^{(n)} \rVert^2.
\end{equation*}
Exploiting the inexact gradient ascent update in Equation~\eqref{eq:update}, we obtain
\begin{equation}
G(\boldsymbol{\theta}^{(n+1)}) \geq G(\boldsymbol{\theta}^{(n)}) + \eta  \nabla_{\boldsymbol{\theta}} G(\boldsymbol{\theta}^{(n)})^\top \hat{g}_n - \dfrac{L\eta^2}{2}\lVert \hat{g}_n \rVert^2.
\label{eq:lsmooth1}
\end{equation}
Taking the expectation of Equation~\eqref{eq:lsmooth1}, we get
\begin{equation*}
G(\boldsymbol{\theta}^{(n+1)}) \geq G(\boldsymbol{\theta}^{(n)}) + \eta\left(1 - \dfrac{L \eta}{2}\right) \lVert \nabla_{\boldsymbol{\theta}} G(\boldsymbol{\theta}^{(n)}) \rVert^2 - \dfrac{L \eta^2}{2} \sigma^2 
\end{equation*}
It is now possible to average over the iterations $n=1,\ldots,N$, exploiting the telescopic sum and obtaining
\begin{equation*}
\dfrac{G(\boldsymbol{\theta}^{(N+1)}) - G(\boldsymbol{\theta}^{(1)})}{N} \geq \eta\left(1-\dfrac{L\eta}{2}\right) \sum_{n=1}^{N} \lVert \nabla_{\boldsymbol{\theta}} G(\boldsymbol{\theta}^{(n)}) \rVert^2 - \dfrac{L\eta^2N}{2}\sigma^2,
\end{equation*}
that is, assuming $\eta < \frac{2}{L}$ and since $G(\boldsymbol{\theta}^*) \geq G(\boldsymbol{\theta}^{(N+1)})$ due to the optimality of $\boldsymbol{\theta}^*$,
\begin{equation*}
\sum_{n=1}^{N} \lVert \nabla_{\boldsymbol{\theta}} G(\boldsymbol{\theta}^{(n)}) \rVert^2 \leq \dfrac{2(G(\boldsymbol{\theta}^{*}) - G(\boldsymbol{\theta}^{(1)}))}{\eta(2-L\eta)N}  +  \dfrac{L\eta^2N}{\eta(2-L\eta)}\sigma^2
\end{equation*}
The learning rate $\eta$ can be selected in order to minimize the right-hand side of the above inequality and achieve a sharper upper bound, even though the same result can be proven also with different step size rules~\cite{Nesterov2015, nesterov2013introductory}. Doing the math, one can obtain
\[
\underset{n=1,\ldots,N}{\min}\lVert \nabla_{\boldsymbol{\theta}} G(\boldsymbol{\theta}^{(n)}) \rVert^2 \leq \dfrac{1}{N}\sum_{n=1}^{N} \lVert \nabla_{\boldsymbol{\theta}} G(\boldsymbol{\theta}^{(n)}) \rVert^2 = 
O\left(\dfrac{\sigma}{\sqrt{N}}\right),
\]
thus guaranteeing the convergence of the algorithm with rate $O\left(\frac{\sigma}{\sqrt{N}}\right)$. In other words, $N=O\left(\frac{\sigma^2}{\varepsilon^2}\right)$ iterations are required to achieve $\underset{n=1,\ldots,N}{\min}\lVert \nabla_{\boldsymbol{\theta}} G(\boldsymbol{\theta}^{(n)}) \rVert^2 \leq \varepsilon$, with $\varepsilon > 0$ being the prescribed tolerance.

\section{Adjoint state method}
\label{sec:appendix2}

In this section, we derive the loss gradient formulas in Equations~\eqref{eq:grad1} and \eqref{eq:grad2}, both in continuous-time and after time discretization. To compute the exact gradient of the loss function $J$ with respect to the policy parameters $\boldsymbol{\theta}$, one may exploit a Hamiltonian formulation, in accordance with the Pontryagin maximum principle~\cite{Pontryagin1962}, a Lagrangian multiplier approach or, as considered in the following, the adjoint state method. Let us consider the OCP in Equation~\eqref{eq:ocpcont}, where the control is determined through the policy $\mathbf{u} = \pi(\mathbf{y},\boldsymbol{\mu}; \boldsymbol{\theta})$. Importantly, the adjoint state method is traditionally applied to directly determine the open-loop optimal control actions, i.e. to compute the gradient $\nabla_\mathbf{u} J$, while we optimize with respect to the policy parameters $\boldsymbol{\theta}$, thus focusing on closed-loop control strategies. The loss gradient $\nabla_{\boldsymbol{\theta}} J$ may be initially computed via the chain rule, yielding
\[
\nabla_{\boldsymbol{\theta}} J =  \int_0^T \left( \dfrac{\partial L}{\partial \mathbf{y}} + \dfrac{\partial L}{\partial \mathbf{u}}\dfrac{\partial \pi}{\partial \mathbf{y}} \right) \dfrac{d \mathbf{y}}{d \boldsymbol{\theta}} + \dfrac{\partial L}{\partial \mathbf{u}}\dfrac{\partial \pi}{\partial \boldsymbol{\theta}} dt + \dfrac{\partial \phi}{\partial \mathbf{y}}(\mathbf{y}(T))\dfrac{d \mathbf{y}}{d \boldsymbol{\theta}}(T),
\]
where $\frac{\partial J}{\partial \boldsymbol{\theta}} = \frac{\partial L}{\partial \boldsymbol{\theta}} = \frac{\partial \phi}{\partial \boldsymbol{\theta}} =  \boldsymbol{0}$ and $\frac{d \boldsymbol{\mu}}{d \boldsymbol{\theta}} = \boldsymbol{0}$. Note that we omit the dependence on $\boldsymbol{\mu}$ for the sake of compactness. Note also that, in order to have a well-defined expression, we assume that all the components -- i.e. the policy $\pi$ and the loss function terms $L$ and $\phi$ are differentiable. The rationale behind the adjoint state method is to introduce an auxiliary time-dependent variable, that is the so-called adjoint or co-state variable $\boldsymbol{\lambda}(t,\boldsymbol{\mu})\in \mathbb{R}^{N_y}$, in order to get rid of the Jacobian $\frac{d \mathbf{y}}{d \boldsymbol{\theta}}$, which is prohibitive to compute. Note that, in general, the semi-discrete adjoint variable may have a different dimension than $N_y$. However, we take into account the same spatial discretization for state and adjoint variables. Let us compute the so-called tangent equation by differentiating the state equation with respect to the policy parameters
\begin{equation}
\begin{cases}
\dfrac{d \dot{\mathbf{y}}}{d \boldsymbol{\theta}} = \left( \dfrac{\partial f}{\partial \mathbf{y}} + \dfrac{\partial f}{\partial \mathbf{u}} \dfrac{\partial \pi}{\partial \mathbf{y}} \right) \dfrac{d \mathbf{y}}{d \boldsymbol{\theta}} + \dfrac{\partial f}{\partial \mathbf{u}}\dfrac{\partial \pi}{\partial \boldsymbol{\theta}}
\\
\dfrac{d \mathbf{y}}{d \boldsymbol{\theta}}(0) = \dfrac{d \mathbf{y}_0}{d \boldsymbol{\theta}} = \boldsymbol{0}
\end{cases}
\label{eq:tangcont}
\end{equation}
where $\frac{\partial f}{\partial \boldsymbol{\theta}} = \boldsymbol{0}$ and $\frac{d \boldsymbol{\mu}}{d \boldsymbol{\theta}} = \boldsymbol{0}$.
It is now possible to add Equation~\eqref{eq:tangcont} multiplied by the negative of the adjoint variable to the time integral of the loss gradient formula above
\[
\nabla_{\boldsymbol{\theta}} J =  \int_0^T \left( \dfrac{\partial L}{\partial \mathbf{y}} + \dfrac{\partial L}{\partial \mathbf{u}}\dfrac{\partial \pi}{\partial \mathbf{y}} \right) \dfrac{d \mathbf{y}}{d \boldsymbol{\theta}} + \dfrac{\partial L}{\partial \mathbf{u}}\dfrac{\partial \pi}{\partial \boldsymbol{\theta}} - \boldsymbol{\lambda}^{\top}
\left(\dfrac{d \dot{\mathbf{y}}}{d \boldsymbol{\theta}} - \left( \dfrac{\partial f}{\partial \mathbf{y}} + \dfrac{\partial f}{\partial \mathbf{u}} \dfrac{\partial \pi}{\partial \mathbf{y}} \right) \dfrac{d \mathbf{y}}{d \boldsymbol{\theta}} - \dfrac{\partial f}{\partial \mathbf{u}}\dfrac{\partial \pi}{\partial \boldsymbol{\theta}} \right)
dt + \dfrac{\partial \phi}{\partial \mathbf{y}}(\mathbf{y}(T))\dfrac{d \mathbf{y}}{d \boldsymbol{\theta}}(T).
\]
After integrating by parts the term $\boldsymbol{\lambda}^{\top}\frac{d \dot{\mathbf{y}}}{d \boldsymbol{\theta}}$, we have
\[
\nabla_{\boldsymbol{\theta}} J =  \int_0^T \left( \dot{\boldsymbol{\lambda}}^{\top} + \boldsymbol{\lambda}^{\top} \left( \dfrac{\partial f}{\partial \mathbf{y}} + \dfrac{\partial f}{\partial \mathbf{u}} \dfrac{\partial \pi}{\partial \mathbf{y}}  \right) + \dfrac{\partial L}{\partial \mathbf{y}} + \dfrac{\partial L}{\partial \mathbf{u}}\dfrac{\partial \pi}{\partial \mathbf{y}} \right) \dfrac{d \mathbf{y}}{d \boldsymbol{\theta}} + \dfrac{\partial L}{\partial \mathbf{u}}\dfrac{\partial \pi}{\partial \boldsymbol{\theta}} + \boldsymbol{\lambda}^{\top}\dfrac{\partial f}{\partial \mathbf{u}}\dfrac{\partial \pi}{\partial \boldsymbol{\theta}} dt + \left(\dfrac{\partial \phi}{\partial \mathbf{y}}(\mathbf{y}(T)) - \boldsymbol{\lambda}^{\top}(T)\right)\dfrac{d \mathbf{y}}{d \boldsymbol{\theta}}(T).
\]
Selecting the adjoint variable as solution of Equation~\eqref{eq:adj1}, we get rid of the term $\frac{d \mathbf{y}}{d \boldsymbol{\theta}}$, and we end up with the loss gradient formula in Equation~\eqref{eq:grad1}. Notice that, in general, one has to consider the conjugate transpose of the matrix and the vector in the adjoint equation.

The same arguments can be applied to the discrete-time optimal control problem in Equation~\eqref{eq:ocpdiscrete} with $\mathbf{u}_k = \pi(\mathbf{y}_k,\boldsymbol{\mu}; \boldsymbol{\theta})$ for $k=0,\ldots,N_t$. Specifically, via chain rule, one can obtain
\[
\nabla_{\boldsymbol{\theta}} J = \sum_{k=0}^{N_t-1} \left[ \left( \dfrac{\partial L}{\partial \mathbf{y}_k} + \dfrac{\partial L}{\partial \mathbf{u}_k} \dfrac{\partial \pi}{\partial \mathbf{y}_k} \right) \dfrac{d \mathbf{y}_k}{d \boldsymbol{\theta}} + \dfrac{\partial L}{\partial \mathbf{u}_k} \dfrac{\partial \pi}{\partial \boldsymbol{\theta}} \right] + \dfrac{\partial \phi}{\partial \mathbf{y}_{N_t}} \dfrac{d \mathbf{y}_{N_t}}{d \boldsymbol{\theta}}.
\]
The tangent equation is now given by
\[
\begin{cases}
 \dfrac{d \mathbf{y}_{k+1}}{d \boldsymbol{\theta}} = \left( \dfrac{\partial F}{\partial \mathbf{y}_k} + \dfrac{\partial F}{\partial \mathbf{u}_k} \dfrac{\partial \pi}{\partial \mathbf{y}_k} \right) \dfrac{d \mathbf{y}_k}{d \boldsymbol{\theta}} + \dfrac{\partial F}{\partial \mathbf{u}_k} \dfrac{\partial \pi}{\partial \boldsymbol{\theta}}
 \\
 \dfrac{d \mathbf{y}_{0}}{d \boldsymbol{\theta}} = \boldsymbol{0}
\end{cases}
\]
for $k=0,\ldots,N_t-1$. We can now add the tangent equation multiplied by the negative of the adjoint variables $\boldsymbol{\lambda}_{k} \in \mathbb{R}^{N_y}$ for $k=0,\ldots,N_t$ in the loss gradient formula above, yielding
\[
\nabla_{\boldsymbol{\theta}} J = \sum_{k=0}^{N_t-1} \left[ \left( \dfrac{\partial L}{\partial \mathbf{y}_k} + \dfrac{\partial L}{\partial \mathbf{u}_k} \dfrac{\partial \pi}{\partial \mathbf{y}_k} \right) \dfrac{d \mathbf{y}_k}{d \boldsymbol{\theta}} + \dfrac{\partial L}{\partial \mathbf{u}_k} \dfrac{\partial \pi}{\partial \boldsymbol{\theta}} - \boldsymbol{\lambda}^\top_{k+1} \left( \dfrac{d \mathbf{y}_{k+1}}{d \boldsymbol{\theta}} - \left( \dfrac{\partial F}{\partial \mathbf{y}_k} + \dfrac{\partial F}{\partial \mathbf{u}_k} \dfrac{\partial \pi}{\partial \mathbf{y}_k} \right) \dfrac{d \mathbf{y}_k}{d \boldsymbol{\theta}} - \dfrac{\partial F}{\partial \mathbf{u}_k} \dfrac{\partial \pi}{\partial \boldsymbol{\theta}}\right)
\right] + \dfrac{\partial \phi}{\partial \mathbf{y}_{N_t}} \dfrac{d \mathbf{y}_{N_t}}{d \boldsymbol{\theta}}.
\]
Rearranging the terms in the sum and exploiting the homogeneous initial condition of the tangent equation above, this is equivalent to 
\[
\nabla_{\boldsymbol{\theta}} J = \sum_{k=0}^{N_t-1} \left[ \left( -\boldsymbol{\lambda}^\top_k +\boldsymbol{\lambda}^\top_{k+1} \left( \dfrac{\partial F}{\partial \mathbf{y}_k} + \dfrac{\partial F}{\partial \mathbf{u}_k} \dfrac{\partial \pi}{\partial \mathbf{y}_k} \right) + \dfrac{\partial L}{\partial \mathbf{y}_k} + \dfrac{\partial L}{\partial \mathbf{u}_k} \dfrac{\partial \pi}{\partial \mathbf{y}_k}  \right) \dfrac{d \mathbf{y}_k}{d \boldsymbol{\theta}} + \dfrac{\partial L}{\partial \mathbf{u}_k} \dfrac{\partial \pi}{\partial \boldsymbol{\theta}} + \boldsymbol{\lambda}^\top_{k+1} \dfrac{\partial F}{\partial \mathbf{u}_k} \dfrac{\partial \pi}{\partial \boldsymbol{\theta}}
\right] + \left(\dfrac{\partial \phi}{\partial \mathbf{y}_{N_t}} - \boldsymbol{\lambda}^\top_{N_t}\right) \dfrac{d \mathbf{y}_{N_t}}{d \boldsymbol{\theta}}.
\]
Selecting the adjoint variables as solutions of Equation~\eqref{eq.adj2}, it is possible to retrieve the loss gradient formula in Equation~\eqref{eq:grad2}.

\section{Equivalence between adjoint state method and reverse-mode automatic differentiation}
\label{sec:appendix3}

\begin{figure*}[t]
    \centering
    \makebox[\linewidth][c]{%
        \includegraphics[width=0.5\linewidth]{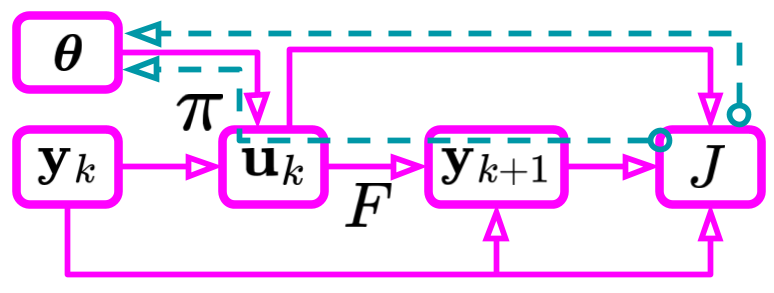}
    }
    \caption{One-step computational graph of a discrete-time optimal control problem. Starting from the state $\mathbf{y}_k$ and the policy parameters $\boldsymbol{\theta}$, it is possible to retrieve the control action $\mathbf{u}_k = \pi(\mathbf{y}_k; \boldsymbol{\theta})$. Moreover, the subsequent state $\mathbf{y}_{k+1}$ is given by the discrete-time transition function, that is $\mathbf{y}_{k+1} = F(\mathbf{y}_k, \mathbf{u}_k)$. Finally, it is possible to compute the loss function, which depends on the states $\mathbf{y}_k,\mathbf{y}_{k+1}$ and the policy parameters $\boldsymbol{\theta}$ through the control action $\mathbf{u}_k$. After the forward pass (solid magenta arrows), it is possible to compute the sensitivities of the loss $J$ to the policy parameters $\boldsymbol{\theta}$ in reverse-mode (dashed teal arrows).}
    \label{fig:compgraph}
\end{figure*}

While the adjoint method provides the mathematical formulas for sensitivities, automatic differentiation provides the software infrastructure to evaluate them efficiently\cite{baydin2018automatic, margossian2019review, 10.1007/3-540-28438-9_2}. Automatic differentiation breaks computer programs into a sequence of simple differentiable operations, and systematically applies the chain rule to retrieve the exact derivatives up to floating-point precision. Specifically, it considers a computational graph where nodes represent variables or intermediate values, while edges represent differentiable operations. Figure~\ref{fig:compgraph} shows the one-step computational graph with reference to the discrete-time optimal control problem in Equation~\ref{eq:ocpdiscrete}, assuming no dependence on scenario parameters $\boldsymbol{\mu}$ without loss of generality.

Differently from forward-mode approaches where derivatives are computed alongside the forward computation, reverse-mode automatic differentiation first stores the variable values through a forward pass, and then propagates gradients backward. This second approach is more efficient when dealing with computational graphs with many inputs and a single output, as typically happens in deep learning, at the cost of a higher memory usage. With reference to the computational graph in Figure~\ref{fig:compgraph}, the gradient $\nabla_{\boldsymbol{\theta}}J$ can be computed by summing the sensitivities of the paths going from $\boldsymbol{\theta}$ to the loss $J$, that are $\boldsymbol{\theta} \to \mathbf{u}_k \to \mathbf{y}_{k+1} \to J$ and $\boldsymbol{\theta} \to \mathbf{u}_k \to J$. In particular, exploiting the chain rule over those paths, one can retrieve 
\[
\nabla_{\boldsymbol{\theta}}J = \dfrac{dJ}{d\boldsymbol{\theta}} = \dfrac{\partial J}{\partial \mathbf{y}_{k+1}} \dfrac{\partial F}{\partial \mathbf{u}_{k}} \dfrac{\partial \pi}{\partial \boldsymbol{\theta}} + \dfrac{\partial J}{\partial \mathbf{u}_{k}} \dfrac{\partial \pi}{\partial \boldsymbol{\theta}},
\]
Similarly, the gradient $\nabla_{\mathbf{y}_k}J$ can be computed backpropagating the derivatives in the graph, that is
\[
\nabla_{\mathbf{y}_k} J = \dfrac{dJ}{d\mathbf{y}_k} = \dfrac{\partial J}{\partial \mathbf{y}_{k+1}} \dfrac{\partial F}{\partial \mathbf{u}_{k}} \dfrac{\partial \pi}{\partial \mathbf{y}_k} + \dfrac{\partial J}{\partial \mathbf{u}_{k}}\dfrac{\partial \pi}{\partial \mathbf{y}_k} + \dfrac{\partial J}{\partial \mathbf{y}_{k+1}}\dfrac{\partial F}{\partial \mathbf{y}_k} + \dfrac{\partial J}{\partial \mathbf{y}_k}.
\]
Notice that, since $J$ depends on the past through the running cost $L$, we have $\frac{\partial J}{\partial \mathbf{u}_k} = \frac{\partial L}{\partial \mathbf{u}_k}$ and $\frac{\partial J}{\partial \mathbf{y}_k} = \frac{\partial L}{\partial \mathbf{y}_k}$. Therefore, if we now define the adjoint variables as the sensitivities $\boldsymbol{\lambda}_k = \frac{dJ}{d\mathbf{y}_k}^{\top} = \nabla_{\mathbf{y}_k} J^\top$ and $\boldsymbol{\lambda}_{k+1} = \frac{dJ}{d\mathbf{y}_{k+1}}^{\top}=\nabla_{\mathbf{y}_{k+1}} J^\top=\frac{\partial J}{\partial \mathbf{y}_{k+1}}^{\top}$, we recover the loss gradient in Equation~\eqref{eq:grad2} and the adjoint equation in Equation~\eqref{eq:adj2}, thus proving the equivalence between the adjoint state method and reverse-mode automatic differentiation, as well as the equivalence in Equation~\eqref{eq:valuegraddiscrete}. The same procedure can be repeated when taking into account multi-step computational graphs, achieving the same result. 

In our setting, reverse-mode automatic differentiation allows us to easily retrieve the gradient of the loss function with respect to the policy parameters. Importantly, differentiable environments are required to propagate the derivatives through the dynamics. To this aim, one may consider numerical physical solvers defined in, e.g., \texttt{PyTorch}~\cite{paszke2017automatic}, \texttt{JAX}~\cite{jax2018github} or \texttt{FEniCS-adjoint}~\cite{mitusch2019dolfin}.

\section{Equivalence between adjoint and value-gradient}
\label{sec:appendix4}
In this section, we show the equivalence between the closed-loop adjoint variable $\boldsymbol{\lambda}(t,\boldsymbol{\mu})$, which satisfies Equation~\eqref{eq:adj1}, and the value gradient $\nabla_\mathbf{y}V_\pi(\mathbf{y},t,\boldsymbol{\mu})$, where the (differentiable and smooth enough, for the sake of argument) value function $V_\pi(\mathbf{y},t,\boldsymbol{\mu})$ is given by Equation~\eqref{eq:valueocp}. Let us compute the total time derivative of the value function using the chain rule
\[
\dfrac{d V_\pi}{dt} = \dot{V_\pi} = \dfrac{\partial V_\pi}{\partial t} + \dfrac{\partial V_\pi}{\partial \mathbf{y}} \dot{\mathbf{y}} = \dfrac{\partial V_\pi}{\partial t} + \dfrac{\partial V_\pi}{\partial \mathbf{y}} f(\mathbf{y}, \pi(\mathbf{y};\boldsymbol{\theta}),\boldsymbol{\mu}),
\]
where $\frac{\partial V_\pi}{\partial \mathbf{y}} =\frac{d V_\pi}{d \mathbf{y}}=\nabla_\mathbf{y}V_\pi$ and
\[
\dfrac{d V_\pi}{dt} = \dfrac{d}{dt}\left[\int_t^T L(\mathbf{y}(\tau),\pi(\mathbf{y}(\tau);\boldsymbol{\theta}),\boldsymbol{\mu})d\tau + \phi(\mathbf{y}(T),\boldsymbol{\mu})\right] = -L(\mathbf{y}(t),\pi(\mathbf{y}(t);\boldsymbol{\theta}),\boldsymbol{\mu})
\]
thanks to the fundamental theorem of calculus. Combining these equations, we obtain the following PDE
\begin{equation}
- \dfrac{\partial V_\pi}{\partial t} = L(\mathbf{y},\pi(\mathbf{y};\boldsymbol{\theta}),\boldsymbol{\mu}) + \nabla_\mathbf{y} V_\pi f(\mathbf{y}, \pi(\mathbf{y};\boldsymbol{\theta}),\boldsymbol{\mu}),
\label{eq:hjbpolicy}
\end{equation}
which corresponds to the Hamilton-Jacobi-Bellman equation equation whenever the optimal policy is taken into account. Differentiating Equation~\eqref{eq:hjbpolicy} with respect to the state, we get
\begin{equation}
- \dfrac{d}{d\mathbf{y}}\dfrac{\partial V_\pi}{\partial t} = -\nabla_\mathbf{y} \dfrac{\partial V_\pi}{\partial t} =\dfrac{\partial L}{\partial \mathbf{y}} + \dfrac{\partial L}{\partial \mathbf{u}}\dfrac{\partial \pi}{\partial \mathbf{y}} + \nabla_\mathbf{y}^2 V_\pi f + \nabla_\mathbf{y} V_\pi \left( \dfrac{\partial f}{\partial \mathbf{y}} + \dfrac{\partial f}{\partial \mathbf{u}}\dfrac{\partial \pi}{\partial \mathbf{y}} \right),
\label{eq:valuegradeq1}
\end{equation}
where $\nabla_\mathbf{y}^2 V_\pi$ is the Hessian matrix of the value function. Notice that, since
\[
\dfrac{d}{d t} \nabla_\mathbf{y} V_\pi = \nabla_\mathbf{y} \dot{V_\pi} = \nabla_\mathbf{y}  \left( \dfrac{\partial V_\pi}{\partial t} + \dfrac{\partial V_\pi}{\partial \mathbf{y}} f \right) = \nabla_\mathbf{y} \dfrac{\partial V_\pi}{\partial t}  + \nabla^2_\mathbf{y} V_\pi f, 
\]
we can rewrite Equation~\eqref{eq:valuegradeq1} as follows
\[
\dfrac{d}{d t} \nabla_\mathbf{y}  V_\pi = - \nabla_\mathbf{y} V_\pi \left( \dfrac{\partial f}{\partial \mathbf{y}} + \dfrac{\partial f}{\partial \mathbf{u}}\dfrac{\partial \pi}{\partial \mathbf{y}} \right) - \left(\dfrac{\partial L}{\partial \mathbf{y}} + \dfrac{\partial L}{\partial \mathbf{u}}\dfrac{\partial \pi}{\partial \mathbf{y}}\right),
\]
that is equivalent to Equation~\eqref{eq:adj1}. Due to the existence and uniqueness of the solutions of these equations, we conclude that $\boldsymbol{\lambda} = \nabla_\mathbf{y}V_\pi^{\top}$. Notably, similar statement can be proven when taking into account the open-loop adjoint variable~\cite{Pontryagin1962} and the value function $
V_\mathbf{u}(\mathbf{y},t,\boldsymbol{\mu}) = \int_t^TL(\mathbf{y}(\tau),\mathbf{u}(\tau),\boldsymbol{\mu})d\tau + \phi(\mathbf{y}(T),\boldsymbol{\mu})$ with no explicit dependence on the parametric policy~\cite{Fleming2006-gv, semeraro, Brunton2022}.

The equivalence between the adjoint variables $\boldsymbol{\lambda}_k(\boldsymbol{\mu}) = \boldsymbol{\lambda}(t_k, \boldsymbol{\mu})$ and the sensitivity of the value function in Equation~\eqref{eq:valueocpdiscrete} to the state $\mathbf{y}_k$ holds also after time discretization. Indeed, differentiating the Bellman equation for a generic policy $\pi$ with respect to the state $\mathbf{y}_k$, one can directly retrieve Equation~\eqref{eq:adj2} requiring that
\[
\boldsymbol{\lambda}_k(\boldsymbol{\mu}) = \nabla_{\mathbf{y}_k}V_\pi(\mathbf{y}_k, \boldsymbol{\mu})^\top.
\]
Moreover, as shown in Appendix~\ref{sec:appendix3}, the discrete adjoint variables are also equal to the loss gradient
\[
\boldsymbol{\lambda}_k = \nabla_{\mathbf{y}_k} J^\top.
\] 
Indeed the loss terms at $t_0,\ldots,t_{k-1}$ vanish when differentiating with respect to $\mathbb{y}_k$ due to time causality.

\end{document}